\documentclass[twocolumn,11pt]{article}
\usepackage{fullpage}
\usepackage[svgnames,dvipsnames,color,table]{xcolor}
\usepackage{hyperref}
\hypersetup{colorlinks,linkcolor={blue},citecolor={blue},urlcolor={RoyalBlue}}

\usepackage{amsthm}
\usepackage{amsmath} 
\usepackage{nicefrac}
\usepackage{bm}
\usepackage{booktabs}
\usepackage{amsfonts}
\usepackage{bbm}
\usepackage{mleftright}
\usepackage{dsfont}
\usepackage{amscd,amssymb,amsmath,amsthm,bbold}
\usepackage{graphicx}
\usepackage{stfloats}
\usepackage{multirow}
\usepackage{wrapfig}
\usepackage[numbers]{natbib}

\usepackage{pifont}
\usepackage{microtype}
\usepackage{enumitem}
\usepackage{xfrac}
\usepackage{parskip}
\usepackage{numprint}
\usepackage{booktabs} 

\usepackage{transparent}

\usepackage{algorithm}
\usepackage{algorithmic}
\usepackage{listings}
\usepackage{xspace}
\usepackage{subcaption}
\usepackage{etoolbox}
\usepackage{comment}
\usepackage{etoc}
\usepackage{wrapfig}
\usepackage{placeins}
\usepackage{makecell}
\usepackage{lipsum,xcolor}

\usepackage{microtype}
\usepackage{booktabs} 
\usepackage{makecell}

\usepackage{amsmath}
\usepackage{amssymb}
\usepackage{mathtools}
\usepackage{amsthm}
\usepackage{chngcntr}
\usepackage{xcolor}

\usepackage{nicefrac}
\usepackage{bm}
\usepackage{booktabs,tabularx}
\usepackage{amsfonts}
\usepackage{bbm}
\usepackage{mleftright}
\usepackage{dsfont}
\usepackage{placeins}
\usepackage{amscd,amssymb,amsmath,amsthm,bbold}
\usepackage[normalem]{ulem}
\usepackage[font=small]{caption}
\usepackage{bbm}
\usepackage{rotating}
\usepackage{xcolor}
\usepackage{tcolorbox}
\usepackage{pgfmath}

\usepackage{epigraph}

\newcommand{\customfootnotetext}[2]{{
  \renewcommand{\thefootnote}{#1}
  \footnotetext[0]{#2}}}

\usepackage{adjustbox}
\usepackage{array}
\usepackage{soul}

\newcolumntype{R}[2]{%
    >{\adjustbox{angle=#1,lap=\width-(#2)}\bgroup}%
    l%
    <{\egroup}%
}

\theoremstyle{plain}

\theoremstyle{definition}

\theoremstyle{remark}

\mathchardef\mhyphen="2D

\usepackage[textsize=tiny]{todonotes}

\begin{document}

\date{}
\title{\Large OpenFlamingo: An Open-Source Framework for Training \\Large Autoregressive Vision-Language Models}

\author{
Anas Awadalla$^{*1}$
\and Irena Gao$^{*2}$
\and Josh Gardner$^{1}$
\and Jack Hessel$^{3}$
\and Yusuf Hanafy$^{1}$
\and Wanrong Zhu$^{5}$
\and Kalyani Marathe$^{1}$
\and Yonatan Bitton$^{6}$
\and Samir Gadre$^{7}$
\and Shiori Sagawa$^{2}$
\and Jenia Jitsev$^{4, 9}$
\and Simon Kornblith$^{8}$
\and Pang Wei Koh$^{1,8}$
\and Gabriel Ilharco$^{1}$
\and Mitchell Wortsman$^{1}$
\and Ludwig Schmidt$^{1,3,4}$ 
}

\maketitle
\customfootnotetext{*}{\scriptsize Equal contribution. $^1$University of Washington $^2$Stanford University $^3$Allen Institute for AI $^4$LAION $^5$University of California Santa Barbara $^6$Hebrew University $^7$Columbia University $^8$Google DeepMind $^9$Juelich Supercomputing Center, Research Center Juelich.
Correspondence to \textless anasa2@cs.washington.edu, irena@cs.stanford.edu, schmidt@cs.washington.edu\textgreater.
}
\newcommand{\eg}{\emph{e.g.}\xspace}
\newcommand{\ie}{\emph{i.e.}\xspace}

\newcommand{\typemodel}{autoregressive vision-language model\xspace} 
\newcommand{\typemodels}{autoregressive vision-language models\xspace}

\newcommand{\of}{{OpenFlamingo}\xspace}
\newcommand{\oftb}{{\of}-3B\xspace}
\newcommand{\offb}{{\of}-4B\xspace}
\newcommand{\ofnb}{{\of}-9B\xspace}

\newcommand{\fl}{Flamingo\xspace}
\newcommand{\fltb}{{\fl}-3B\xspace}
\newcommand{\flnb}{{\fl}-9B\xspace}

\newcommand{\kosmos}{Kosmos-1\xspace}
\newcommand{\coco}{COCO\xspace}
\newcommand{\vqav}{VQAv2\xspace}
\newcommand{\flickr}{Flickr-30K\xspace}
\newcommand{\okvqa}{OK-VQA\xspace}
\newcommand{\textvqa}{TextVQA\xspace}
\newcommand{\vizwiz}{VizWiz\xspace}
\newcommand{\hms}{HatefulMemes\xspace}

\newcommand{\image}{{\small\texttt{<image>}}\xspace}
\newcommand{\eoc}{{\small\texttt{<|endofchunk|>}}\xspace}
\newcommand{\gain}[2]{%
    \pgfmathsetmacro{\diff}{#1 - #2}%
    \ifdim\diff pt<0pt
        \textcolor{red}{(\pgfmathprintnumber[fixed,precision=1]{\diff})}%
    \else
        \textcolor{ForestGreen}{(+\pgfmathprintnumber[fixed,precision=1]{\diff})}%
    \fi
}

\definecolor{lightgrey}{rgb}{0.95, 0.95, 0.95}
\sethlcolor{lightgrey}

\newif\ifcomments
\commentstrue

\ifcomments
    \newcommand{\irena}[1]{\textcolor{purple}{[Irena: #1]}}
    \newcommand{\anas}[1]{\textcolor{pink}{[Anas: #1]}}
    \newcommand{\pw}[1]{\textcolor{ForestGreen}{[PW: #1]}}
    \newcommand{\zwr}[1]{\textcolor{blue}{[Wanrong: #1]}}
    \newcommand{\jg}[1]{\textcolor{blue}{[Josh: #1]}}
    \newcommand{\jack}[1]{\textcolor{red}{[Jack: #1]}}
    \newcommand{\yonatan}[1]{\textcolor{blue}{[Yonatan: #1]}}
    \newcommand{\todorices}[1]{\textcolor{red}{[Revise to random instead of RICES] #1}}
\else
    \newcommand{\irena}[1]{\textcolor{purple}{}}
    \newcommand{\anas}[1]{\textcolor{green}{}}
    \newcommand{\pw}[1]{\textcolor{ForestGreen}{}}
    \newcommand{\zwr}[1]{\textcolor{blue}{}}
    \newcommand{\jg}[1]{\textcolor{blue}{}}
    \newcommand{\jack}[1]{\textcolor{red}{}}
    \newcommand{\yonatan}[1]{\textcolor{blue}{}}
    \newcommand{\todorices}[1]{#1}
\fi

\thispagestyle{empty}

\begin{abstract}
We introduce \of, a family of \typemodels ranging from 3B to 9B parameters.
\of is an ongoing effort to produce an open-source replication of DeepMind's \fl models~\cite{alayrac2022flamingo}.
On seven vision-language datasets, 
\of models average between 80 - 89\% of corresponding \fl performance.
This technical report describes our models, training data, hyperparameters, and evaluation suite. 
We share our models and code at \url{https://github.com/mlfoundations/open_flamingo}.
\end{abstract}

\section{Introduction}

\begin{figure}[thb]
    \centering
    \includegraphics[width=\linewidth]{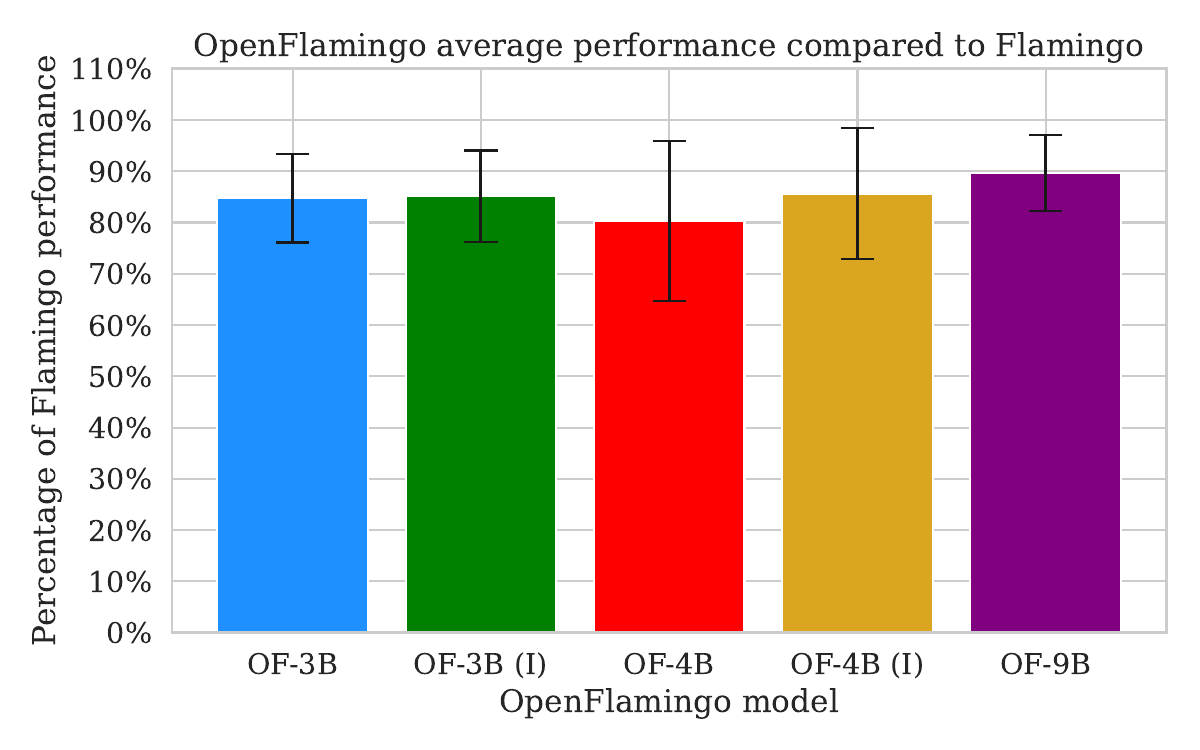}
    \caption{\of performance as a fraction of corresponding \fl performance, averaged across evaluation settings (7 datasets $\times$ 5 options for number of in-context examples). Demonstrations are chosen using RICES (Retrieval-based In-Context Example Selection). More details regarding selecting demonstrations can be found in Section~\ref{sec:eval_method}. We compare \of-3B and -4B models to \fl-3B, and \ofnb to \fl-9B. Error bars are standard deviations over settings. ``OF-3B (I)'' refers to \oftb (Instruct), the 3B model trained with a language-instruction-tuned backbone.
    \vspace{-0em}
    }
    \label{fig:relative_avg_across_all}
\end{figure}

\begin{figure*}[thb]
    \centering
    \includegraphics[width=\linewidth]{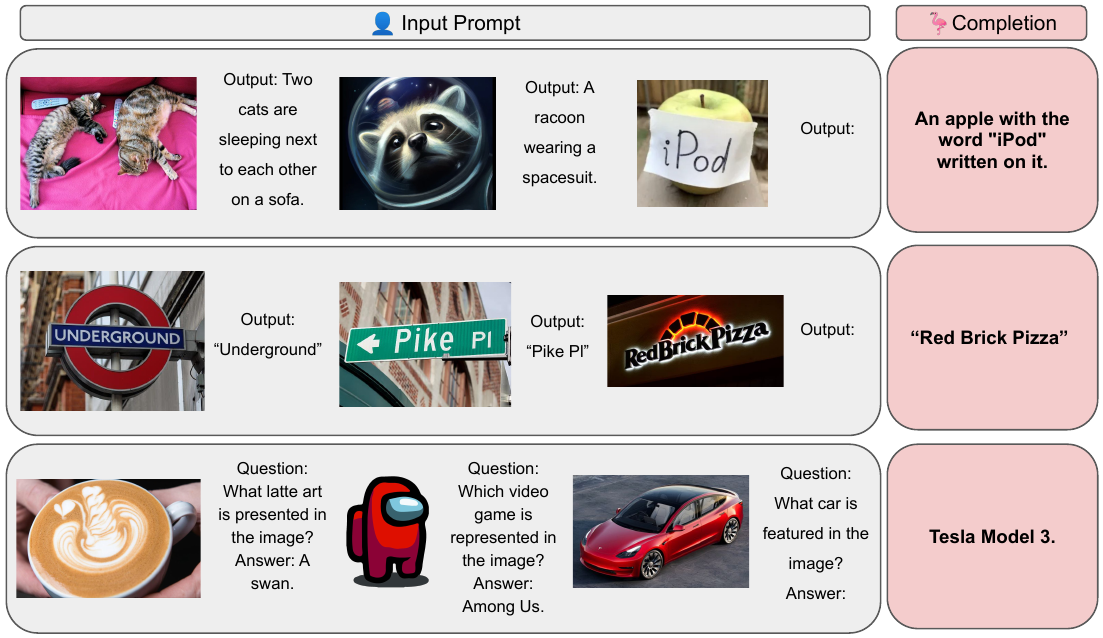}
    \caption{\ofnb (pictured) can process interleaved image-and-text sequences. This interface allows \of to learn many vision-language tasks through in-context demonstrations.\vspace{-0em}
    }    
    \label{fig:samples}
\end{figure*}


A popular format for vision and language models is (image, text) $\rightarrow$ text, i.e., models take as input an image and some text, and produce text as output, e.g., BLIP-2 \cite{li2022blip}.
The flexible format directly 
supports tasks like image classification
and visual question answering (VQA).

However, assuming a single image as input is limiting:
\typemodels enable new capabilities by instead mapping an arbitrarily interleaved \emph{sequence} of images and text to textual outputs.
This interface provides important flexibility:
the input sequence can include demonstrations for a new task, enabling few-shot, in-context learning \cite{alayrac2022flamingo} or multi-round multi-modal chatbot interactions.
Evaluations suggest that \typemodels can be performant foundation models \cite{bommasani2021opportunities}:
models like \fl \cite{alayrac2022flamingo}, CM3 \cite{Aghajanyan2022CM3AC}, \kosmos \cite{huang2023kosmos}, PALM-E~\cite{Driess2023PaLMEAE}, and multimodal GPT-4 \cite{openai2023gpt} generalize well across diverse vision-language tasks.

Unfortunately, these \typemodels are closed-source, and their weights, training data, code, and hyperparameters are proprietary.
This limits the academic community's ability to conduct research on \typemodels, e.g., to understand how web-scraped image-text data affects models' performance and safety.
Open-source alternatives, such as LLaVA \cite{liu2023visual}, LLaMA-Adapter \cite{zhang2023llama}, BLIP-2~\cite{Li2023BLIP2BL}, and mPLUG-Owl \cite{ye2023mplug}, only take in single images, and they often directly train on curated datasets like \coco \cite{lin2014microsoft} rather than web data.

In this technical report, we document our experiences building an open-source reproduction of the \fl models \cite{alayrac2022flamingo}. 
Following \fl, we augment the layers of pretrained, frozen language models so
that they cross attend to the outputs of a frozen
vision encoder while predicting the next token.
The cross-modal module is trained on web-scraped image-text sequences, in our case, two open source datasets: LAION-2B \cite{schuhmann2022laion} and Multimodal C4 \cite{zhu2023multimodal}.
Our stack is built using publicly available components, including CLIP as a vision encoder \cite{radford2021learning} and open-source language models as decoders \cite{mpt,redpajama3b}.

We call the resulting family of five models \of. 
These models range from 3B to 9B parameters, with both standard and instruction-tuned \cite{wei2021finetuned} language model backbones.
When averaging performance across 7 evaluation datasets, \of-3B and -9B models attain 85\% and 89\% of their corresponding Flamingo models respectively (Figure \ref{fig:relative_avg_across_all}).
Models and code are open-sourced at \url{https://github.com/mlfoundations/open_flamingo}.

\begin{table*}[tbhp]
  \centering
  \small 
  \caption{Architecture details of the \of models. All five models use a CLIP ViT-L/14 vision encoder \cite{radford2021learning}. A cross-attention interval of $4$ means that a cross-attention module is inserted every 4th language model layer. Note that \of models labeled {(Instruct)} use language models that were finetuned on language-only tasks; we have not instruction-tuned \of models on vision-language tasks.}
  \begin{tabularx}{\textwidth}{l|l|X|X} 
    \toprule
    \textbf{Model}  & \textbf{Language model} & \textbf{Cross-attention interval} & \textbf{\image and \eoc} \\
    \midrule
    {\oftb}   & MPT-1B \cite{mpt} &  1   & Trainable \\
    {\oftb (Instruct)}     &  MPT-1B (Instruct) \cite{mpt} &  1   & Trainable \\
    {\offb}   &  RedPajama-3B \cite{redpajama3b} &  2   & Frozen \\
    {\offb (Instruct)}   &  RedPajama-3B (Instruct) \cite{redpajama3b} &  2   & Frozen \\
    {\ofnb}   &  MPT-7B \cite{mpt} &  4   & Trainable \\
    \bottomrule
  \end{tabularx}%
  \label{tab:models}%
\end{table*}%

\section{Related work}
\textbf{Generative vision-language models} output text conditioned on an image-text sequence.
While many such architectures, such as BLIP-2 and LLaVa, can incorporate only one image in their context \cite{chen2022pali,koh2023grounding,li2022blip,liu2023visual,ye2023mplug,zhang2023llama}, \typemodels accept interleaved image-text sequences, enabling in-context learning.

We chose to replicate \fl because of its strong in-context learning abilities. Aggregated across evaluation sets, \fl models see steady performance improvements up to 32 in-context examples \cite{alayrac2022flamingo}.
This is in contrast with other \typemodels, for example \kosmos \cite{huang2023kosmos}; on captioning tasks \coco \cite{lin2014microsoft} and \flickr \cite{plummer2015flickr30k}, \kosmos shows performance improvements up to 4 in-context examples, but performance degrades when using 8 in-context examples.


\paragraph{Open-source image-text datasets.}
Proprietary \typemodels are typically trained on closed-source datasets \cite{Aghajanyan2022CM3AC,alayrac2022flamingo,Driess2023PaLMEAE,huang2023kosmos}. 
For example, \fl relies on image-text pairs from the ALIGN dataset \cite{jia2021align} and interleaved image-text sequences from the M3W dataset \cite{alayrac2022flamingo}; both are unavailable to the public.
Recent efforts to replicate these web-scraped datasets include LAION-2B, a dataset of image-text pairs, and Multimodal C4 \cite{zhu2023multimodal} and OBELISC \cite{obelisc}, datasets of image-text sequences.
We use LAION-2B and Multimodal C4 for training \of models.
\citet{obelisc} also train 9B and 80B \fl-style models; their models differ in the choice of pretraining dataset (OBELISC instead of Multimodal C4) and language model (LLaMA-9B \cite{zhang2023llama} instead of the MPT and RedPajama-3B models \cite{mpt,redpajama3b}).

\section{Approach}
\subsection{Architecture}\label{sec:architecture}

We match the \fl architecture \cite{alayrac2022flamingo}.
Given an interleaved sequence of images with text tokens, \of models predict the next text token conditioned on all previous text tokens and the last preceding image. 
Text tokens attend to their corresponding images via \emph{dense cross-attention modules}, which we attach to the layers of a frozen, autoregressive language model.
To embed images, we extract patch features from a frozen vision encoder and pass these through a trainable Perceiver resampler \cite{jaegle2021perceiver}. 

As a preprocessing step, we first mark the locations of images in the text sequence with \image tokens.
We also insert \eoc tokens after the text tokens following an image; \eg the sequence \hl{$x$ Hello world}, where $x$ is an image, would be preprocessed into \hl{\image~Hello world \eoc~}.

Unlike \fl, we do not support video inputs at this time. We leave this for future work.

Table \ref{tab:models} describes the five \of models based on their language model and density of cross-attention layers; all models use CLIP ViT-L/14 \cite{radford2021learning} as a vision encoder.
In most cases, the \image and \eoc embeddings are trainable, while other text embeddings are frozen.
For the \offb models, all embeddings are frozen, including the randomly initialized \image and \eoc embeddings.
This was due to complications with gradient masking when using Fully Sharded Data Parallel (\S\ref{sec:distributed}).

\subsection{Training data}\label{sec:data}
We train our models on a mixture of image-text pairs and interleaved image-text sequences. 
During training, we sample dataset shards with replacement using the WebDataset format \cite{webdataset}.

\begin{table}[tbp]
  \centering
  \small
  \caption{Statistics for training datasets. ``ChatGPT'' stands for the ChatGPT-generated sequences. The median numbers of images and tokens per sequence were calculated using a random sample of 1,000 sequences.}
  \begin{tabularx}{\linewidth}{lXX}
    \toprule
    \textbf{Dataset}& \textbf{Median images per sequence} & \textbf{Median tokens per sequence}\\
    \midrule
    LAION-2B  & 1 & 17 \\
    MMC4 & 2 & 256 \\
    ChatGPT & 3 & 56 \\
    \bottomrule
  \end{tabularx}
    \label{tab:datasets}
\end{table}

\begin{figure*}[t] 
  \centering
  \includegraphics[width=\linewidth]{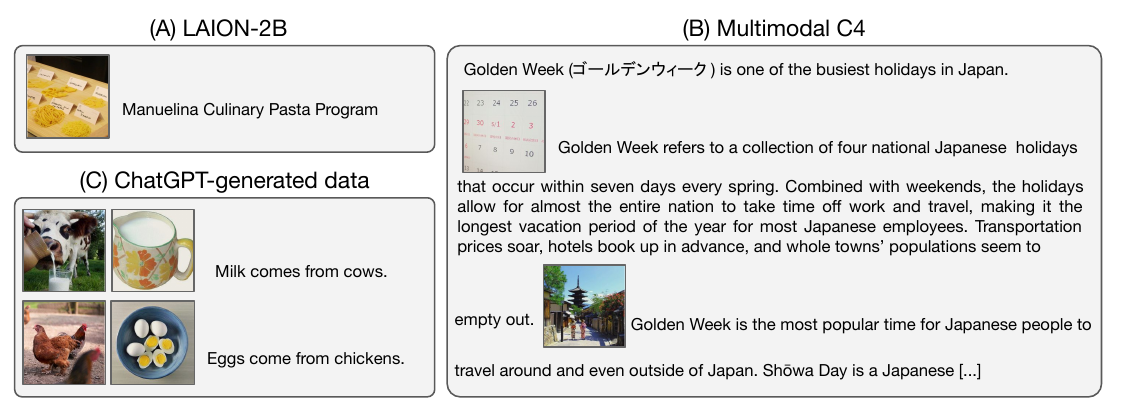} 
  \caption{Samples from (A) LAION-2B \cite{schuhmann2022laion}, (B) Multimodal C4 \cite{zhu2023multimodal}, and (C) ChatGPT-generated data.}
  \label{fig:datasets}
\end{figure*}

\paragraph{LAION-2B \cite{schuhmann2022laion}.} When training \fl, \citet{alayrac2022flamingo} use ALIGN \cite{jia2021align}, a closed-source dataset of over 1B single images paired with short alt-text captions.
To train \of, we replace ALIGN with LAION-2B, an open-source web-scraped dataset consisting of 2B image-text pairs (Figure \ref{fig:datasets}A).
We use part of the English subset and truncate captions to 32 tokens.
All image-text pairs in LAION-2B have a cosine similarity of at least 0.28 according to CLIP ViT-B/32.

\begin{figure}[t] 
  \centering
  \includegraphics[width=\linewidth]{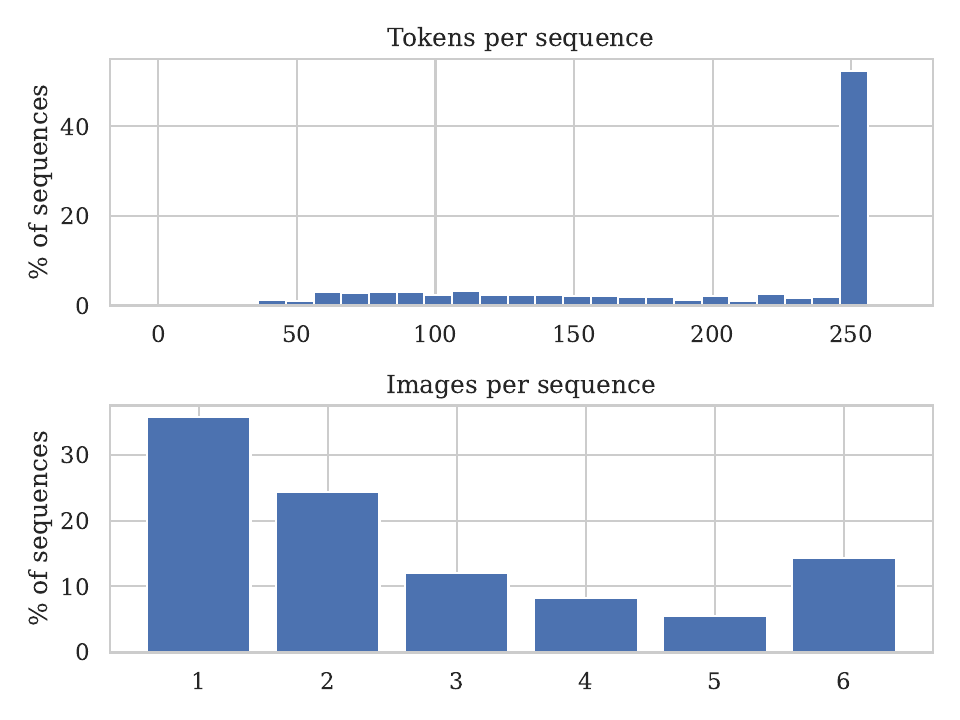} 
  \caption{Histograms of the number of text tokens and images per MMC4 sequence, based on a sample of 1,000 sequences. Sequences are long with few images.\vspace{-0em}}
  \label{fig:mmc4_hist}
\end{figure}

\paragraph{Multimodal C4 \cite{zhu2023multimodal}.}
In addition to image-text pairs, \citet{alayrac2022flamingo} train \fl using M3W, an internal web-scraped dataset of 43M interleaved image-text sequences.
We replace M3W with Multimodal C4 (MMC4), an open-source dataset of 101M interleaved samples  (Figure \ref{fig:datasets}B).
Unlike M3W or OBELISC \cite{obelisc}, which directly parse HTML documents to extract multimodal sequences, MMC4 uses CLIP to soft align images with sentences in a document.
To ensure data quality, we exclude images if their cosine similarity with the subsequent text falls below 0.24, according to CLIP ViT-L/14.
Sequences contain between 1 and 6 images (median 2).
To encourage learning from sequences with multiple images, we reject single-image sequences with probability $0.5$. 
The resulting distribution is shown in Figure \ref{fig:mmc4_hist}.
Additional notes on MMC4 filtering are in Appendix \ref{app:mmc4}.

\label{sec:sec_with_synthetic_data}
\paragraph{Synthetic data.} For the \offb models, we also experimented with training on ChatGPT-generated synthetic data (Figure \ref{fig:datasets}C)
These 417K image-text sequences were generated by prompting ChatGPT to generate a sequence of interleaved text and image alt-texts (in place of images). The alt-texts are used to retrieve a corresponding images from LAION-5B. Additional details of the prompting and data construction process are described in Appendix \ref{sec:sec_with_synthetic_prompt}.
The median number of images per sequence is higher than in MMC4, while the median number of text tokens is lower (Table \ref{tab:datasets}). 
We release these sequences through the \href{https://github.com/mlfoundations/open_flamingo/}{\of repository.}


\subsection{Training details}
\of models were trained for 60M interleaved (MMC4) examples\footnote{\offb models use both MMC4 and ChatGPT-generated data as interleaved sequences; 60M interleaved examples translates to approximately 240K ChatGPT-generated sequences and 59.8M MMC4 sequences. Other models train on 60M MMC4 examples.} and 120M LAION-2B examples.
All models are trained using the next-token prediction objective and optimized with AdamW.
The learning rate is linearly increased at the beginning of training, and then held constant at 1e-4 throughout training.
We apply weight decay of 0.1 on the dense cross attention layers.
The batch size for LAION-2B is twice the batch size of the interleaved dataset (MMC4, optionally with ChatGPT-generated sequences), and the loss weights are set to \fl defaults of $1$ and $0.2$ for MMC4 and LAION-2B respectively.
We accumulate gradients over both datasets between optimizer steps.

\paragraph{Distributed training.}\label{sec:distributed}
We train all models using 64 GPUs distributed across 8 nodes on Stabilty AI's cluster (Table \ref{tab:distributed}).
\offb models were trained using model sharding with Fully Sharded Data Parallel \cite{zhao2023fsdp}; other models were trained using only data parallel.

\begin{figure}[tbh]
\centering
\includegraphics[width=0.41\textwidth]{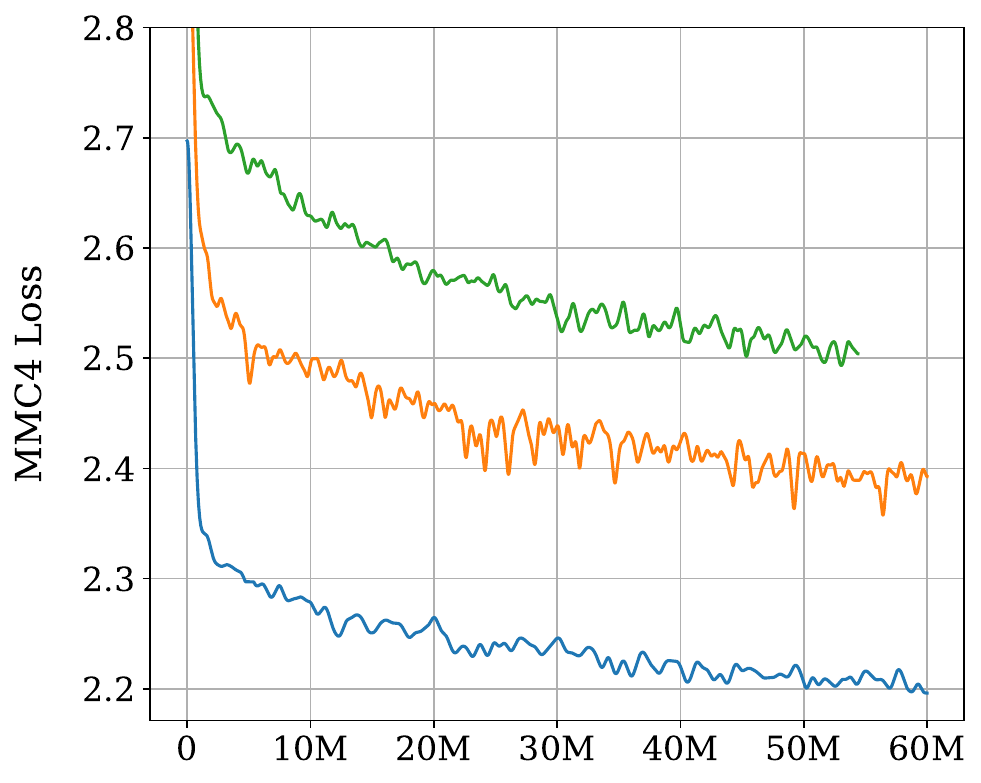}
\includegraphics[width=0.4\textwidth]{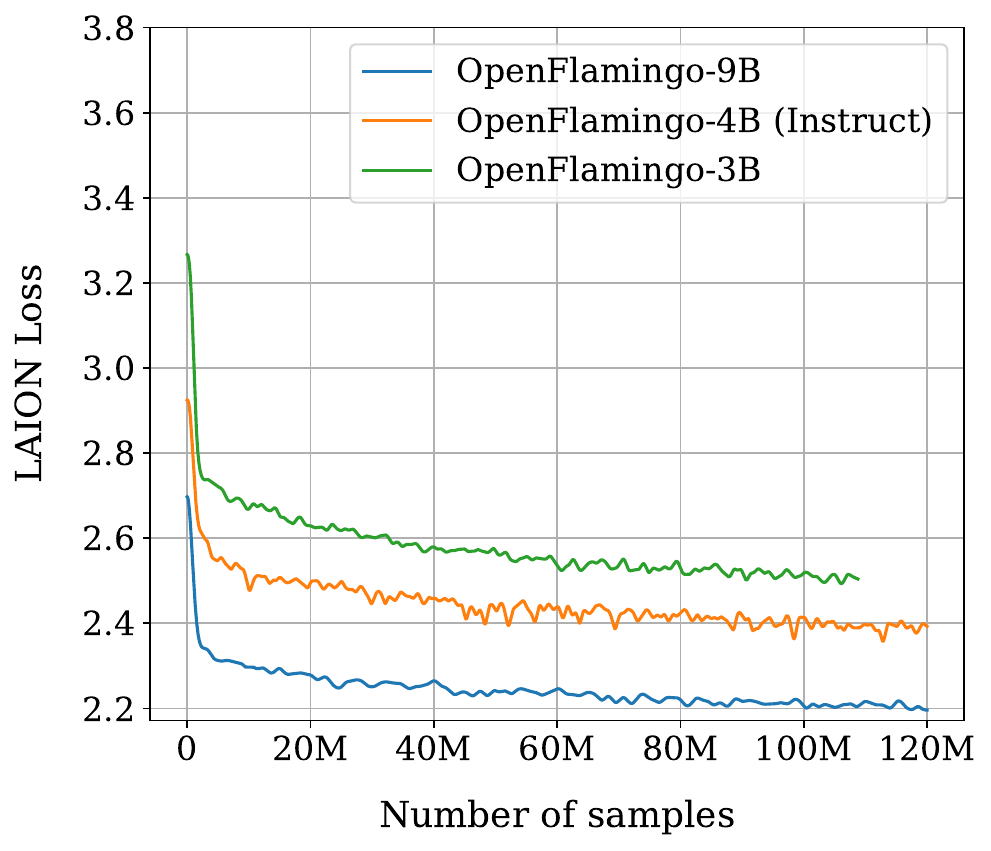}
\caption{MMC4 and LAION-2B language modeling loss throughout training. Curves shown with Gaussian smoothing with window size 100.\vspace{-0em}}
\label{fig:loss}
\end{figure}

\paragraph{Loss curves.} Figure \ref{fig:loss} tracks LAION-2B and MMC4 loss over the course of training. 
After an initial improvement, MMC4 loss decreases very slowly.
We speculate that, since MMC4 sequences tend to include long paragraphs between images (Figure \ref{tab:datasets}), most text tokens can be generated without referencing the image.
Thus, the loss may be dominated by whether the frozen language model can fit unrelated paragraphs of text.

\begin{table}[tbp]
  \centering
  \small
  \caption{Training used either DistributedDataParallel (DDP) or FullyShardedDataParallel (FSDP) \cite{zhao2023fsdp}.}
  \begin{tabularx}{\linewidth}{llXX}
    \toprule
    \textbf{Model} & \textbf{GPU type} & \textbf{Sharding strategy} & \textbf{Precision} \\
    \midrule
    OF-3B &A100-80GB & DDP & fp32 \\
    OF-3B (I) & A100-40GB & DDP & fp32\\
    OF-4B & A100-40GB & FSDP& fp32 \\
    OF-4B (I) & A100-40GB & FSDP & fp32\\
    OF-9B&A100-80GB & DDP & amp\_bf16\\
    \bottomrule
  \end{tabularx}
  \label{tab:distributed}
\end{table}

\subsection{Evaluation method}\label{sec:eval_method}
We evaluate \of on seven vision-language datasets including captioning (\coco \cite{Chen2015MicrosoftCC}, \flickr \cite{Young2014FromID}), visual question answering (\vqav \cite{Agrawal2015VQAVQ}, \okvqa \cite{Marino2019OKVQAAV}, \textvqa \cite{Singh2019TowardsVM}, \vizwiz \cite{Gurari2018VizWizGC}), and rank classification (\hms \cite{Kiela2020TheHM}). 
For each dataset, we measure performance at 0, 4, 8, 16, and 32 in-context examples.
Evaluation was done in automatic mixed precision, with linear layers computed in bfloat16.

\paragraph{Selecting in-context examples.}
For each evaluation example, we sample in-context examples from the training split uniformly at random. 
Additionally, in Appendix \ref{app:rices}, we include evaluations of \of using Retrieval-based In-Context Example Selection (RICES) \cite{yang2022empirical}.

\paragraph{Evaluation subsets.}
We evaluate on the dataset splits used by \citet{alayrac2022flamingo}.
We run each evaluation across three seeds, where the randomness is over selected in-context demonstrations, and average the results to obtain our final scores.

\paragraph{Prompts.}
For captioning tasks, we format demonstrations as \hl{\image~Output: {\small\texttt{[caption]}}}, replacing {\small\texttt{[caption]}} with the ground-truth caption. For VQA, we format examples as \hl{\image~Question: {\small\texttt{[question]}} Short answer: {\small\texttt{[answer]}}}. For \hms, we prompt the model with \hl{\image~is an image with: `{\small\texttt{[text]}}' written on it. Is it hateful? Answer: {\small\texttt{[answer]}}}.

Following \citet{alayrac2022flamingo}, we prompt the model with two in-context examples during zero-shot evaluations, removing their images, and for classification tasks, we implement prompt ensembling by averaging logits across 6 permutations of the in-context examples.

\paragraph{Decoding parameters.} 
We evaluate captioning and VQA using beam search with 3 beams, stopping generation at 20 tokens for captioning, 5 tokens for VQA, or whenever the model produces an \eoc token. 
For \hms, we compute the log-likelihood of completions ``yes'' and ``no'' and answer with the most likely completion.

\paragraph{Metrics.}
For captioning, we use CIDEr score~\cite{Vedantam2014CIDErCI}.
For VQA, we report VQA accuracy, \ie, exact match accuracy over a set of ground truth answers \cite{Agrawal2015VQAVQ}.
For \hms, we compute AUC ROC. 

\section{Results}\label{sec:results}

\begin{figure*}[t] 
  \centering
  \includegraphics[width=\linewidth]{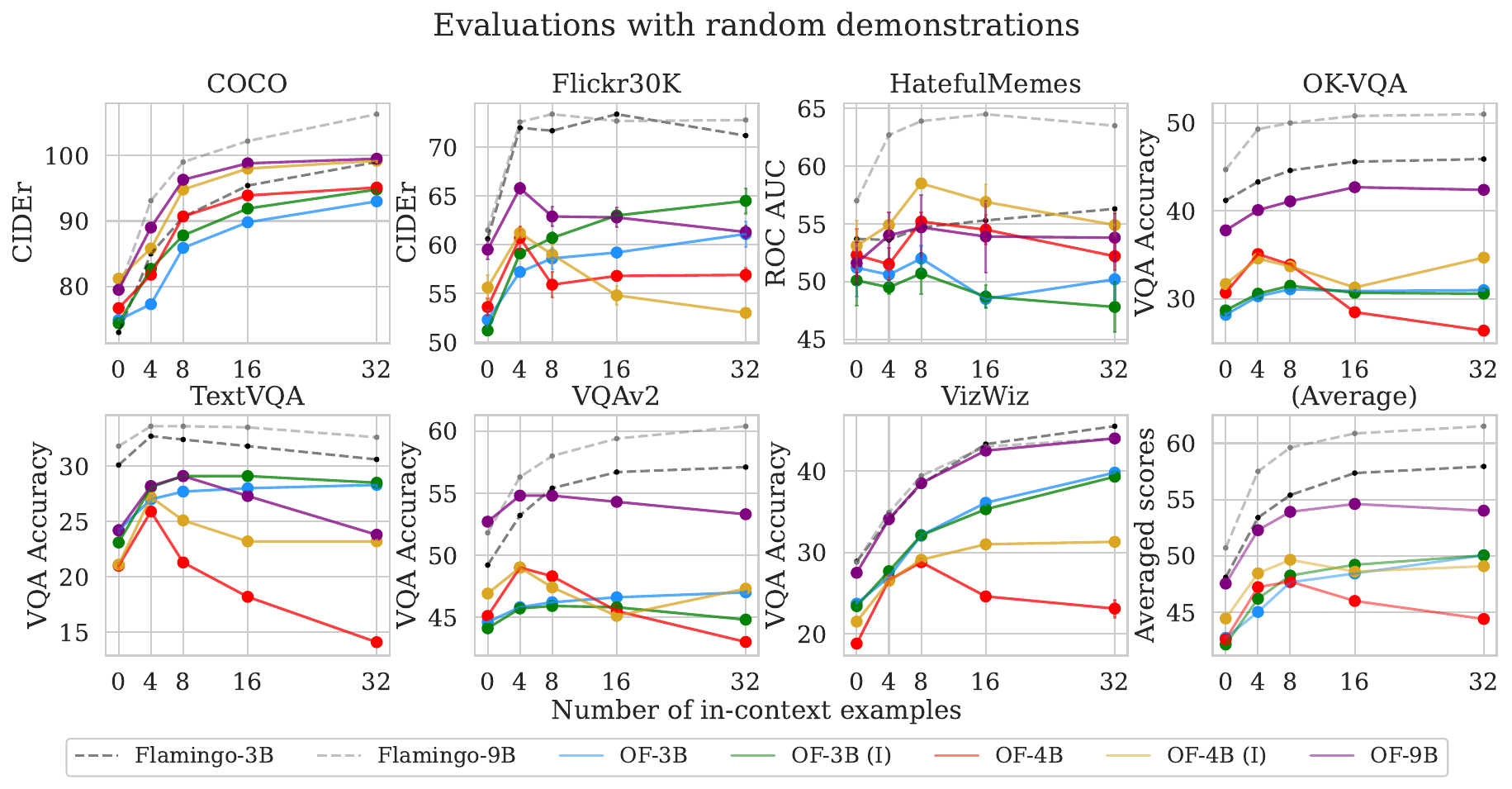} 
  \caption{Evaluation results per dataset across 0, 4, 8, 16, and 32 in-context examples. Each point is the average across three evaluation runs, where the randomness is over choice of in-context demonstrations. Error bars are standard deviations over random seeds. Results are reported in tabular form in Table \ref{tab:eval_large_norices}.\vspace{-0em}}
  \label{fig:results}
\end{figure*}

In Table \ref{tab:eval_short_rices}, we compare \of and \fl models across 0, 4, and 32 in-context examples.
On average, \oftb, -3B (Instruct), -4B (Instruct), and -9B attain more than 86\% of the performance of their corresponding Flamingo models (Figure \ref{fig:relative_avg_across_all}).

In the 0- and 4-shot regimes, \of models approach or match \fl performances on several datasets.
For example, \ofnb improves upon \fl-9B's 0-shot performance on \vqav ($51.8\% \rightarrow 52.7\%$ VQA accuracy) and \coco ($79.4 \rightarrow 79.5$ CIDEr), 
and \ofnb approaches \fl-9B's 0-shot performance on \flickr and \vizwiz. Moreover, \ofnb approaches the 4-shot performance of \fl-9B on \coco, \vqav, and \vizwiz.

However, on \okvqa and \textvqa, \of models are notably weaker than their \fl counterparts: \ofnb underperforms \fl-9B in 0-shot evaluations by 6.9 percentage points on \okvqa and 7.8 percentage points on \textvqa.
\oftb also underperforms \fl-3B by 4.6 percentage points in 0-shot \vqav accuracy.
The reason for generally low VQA performance is unclear, although discussions in \S\ref{sec:vqa} may be related.

\paragraph{Extrapolating to more in-context examples.} 
In Figure \ref{fig:results}, we plot performance as a function of the number of in-context examples.
We observe that the \oftb and -9B models generally improve with the number of in-context examples.
However, the rate of improvement is lower than the \fl models: in the bottom right corner of Figure \ref{fig:results}, we observe that gaps between \ofnb and \fl-9B widen with the number of in-context examples.
We speculate that this behavior may stem from the quality of our pre-training data, which mostly consists of sequences with few images (Table \ref{tab:datasets}).
In contrast with the -3B and -9B models, which generally improve with more in-context examples, the \offb models unexpectedly degrade in performance after 4 or 8 shots.
The 4B models use RedPajama language models \cite{redpajama3b} instead of MPT backbones \cite{mpt}; they also use frozen \image and \eoc embeddings. 
We investigate the effect of the latter in \S\ref{sec:frozen}.




\paragraph{Trends by model size.} 
\ofnb generally outperforms smaller models, except on \hms and for large numbers of in-context examples on \flickr and \textvqa.
However, \offb models often underperform the smaller 3B models, including on \flickr, \hms, \textvqa, and \vizwiz. 

\begin{table*}[ht]
\centering
\resizebox{\textwidth}{!}{
\begin{tabular}{l|c||c|c||c|c|c|c|c}
\textbf{Benchmark} & \textbf{Shots} & \textbf{Fl-3B} & \textbf{Fl-9B} & \textbf{OF-3B} & \textbf{OF-3B (I)} & \textbf{OF-4B} & \textbf{OF-4B (I)} & \textbf{OF-9B}\\ 
\hline
\multirow{5}{*}{\textbf{\coco} \cite{Chen2015MicrosoftCC}} & 0  & 73.0 & 79.4 & 74.9 (0.2) & 74.4 (0.6) & 76.7 (0.2) & \textbf{81.2 (0.3)} & 79.5 (0.2)\\
& 4  & 85.0 & \textbf{93.1} & 77.3 (0.3) & 82.7 (0.7) & 81.8 (0.4) & 85.8 (0.5) & 89.0 (0.3)\\
& 32 & 99.0 & \textbf{106.3} &  93.0 (0.6) & 94.8 (0.3) & 95.1 (0.3) & 99.2 (0.3) & 99.5 (0.1)\\
\hline
\multirow{5}{*}{\textbf{\flickr} \cite{Young2014FromID}} & 0 & 60.6 & \textbf{61.5} & 52.3 (1.0) & 51.2 (0.2) & 53.6 (0.9) & 55.6 (1.3) & 59.5 (1.0)\\
& 4 & 72.0 & \textbf{72.6} & 57.2 (0.4) & 59.1 (0.3) & 60.7 (1.2) & 61.2 (0.5) & 65.8 (0.6)\\
& 32 & 71.2 & \textbf{72.8} &  61.1 (1.3) & 64.5 (1.3) & 56.9 (0.7) & 53.0 (0.5) & 61.3 (0.7)\\
\hline
\multirow{5}{*}{\textbf{\vqav} \cite{Agrawal2015VQAVQ}} 
& 0 & 49.2 & 51.8 &  44.6 (0.0) & 44.1 (0.1) & 45.1 (0.1) & 46.9 (0.0) & \textbf{52.7 (0.2)}\\
& 4 & 53.2 & \textbf{56.3} & 45.8 (0.0) & 45.7 (0.1) & 49.0 (0.0) & 49.0 (0.0) & 54.8 (0.0)\\
& 32 & 57.1 & \textbf{60.4} & 47.0 (0.1) & 44.8 (0.1) & 43.0 (0.2) & 47.3 (0.0) & 53.3 (0.1)\\
\hline
\multirow{5}{*}{\textbf{\okvqa} \cite{Marino2019OKVQAAV}} 
& 0 & 41.2 & \textbf{44.7} & 28.2 (0.2) & 28.7 (0.1) & 30.7 (0.1) & 31.7 (0.1) & 37.8 (0.2)\\
& 4 & 43.3 & \textbf{49.3} & 30.3 (0.5) & 30.6 (0.2) & 35.1 (0.0) & 34.6 (0.0) & 40.1 (0.1)\\
& 32  & 45.9 & \textbf{51.0} & 31.0 (0.1) & 30.6 (0.1) & 26.4 (0.2) & 34.7 (0.3) & 42.4 (0.0)\\
\hline
\multirow{5}{*}{\textbf{\textvqa} \cite{Singh2019TowardsVM}} & 0 & 30.1 & \textbf{31.8} & 24.2 (0.2) & 23.1 (0.2) & 21.0 (0.3) & 21.1 (0.4) & 24.2 (0.5)\\
& 4  & 32.7 & \textbf{33.6} &  27.0 (0.3) & 28.1 (0.4) & 25.9 (0.0) & 27.2 (0.3)  & 28.2 (0.4)\\
& 32 & 30.6 & \textbf{32.6} & 28.3 (0.2) & 28.5 (0.1) & 14.1 (0.2) & 23.2 (0.2) & 23.8 (0.2)\\
\hline
\multirow{5}{*}{\textbf{\vizwiz} \cite{Gurari2018VizWizGC}} 
& 0 & \textbf{28.9} & 28.8 & 23.7 (0.5)  & 23.4 (0.3) & 18.8 (0.1) & 21.5 (0.2) & 27.5 (0.2)\\
& 4 & 34.0 & \textbf{34.9} &27.0 (0.3) & 27.7 (0.1) & 26.6  (0.5)& 26.5 (0.4) & 34.1 (0.7)\\
& 32 & \textbf{45.5} & 44.0 & 39.8 (0.1) & 39.3 (0.4) & 23.1 (1.1) & 31.3 (0.2) & 44.0 (0.5)\\
\hline
\multirow{5}{*}{\textbf{\hms} \cite{Kiela2020TheHM}} & 0 & 53.7 & \textbf{57.0} & 51.2 (2.5) & 50.1 (2.2) & 52.3 (2.3) & 53.1 (2.2) & 51.6 (1.8) \\
& 4 & 53.6 & \textbf{62.7} & 50.6 (0.8) & 49.5 (0.6) & 51.5 (1.4) & 54.9 (1.1) & 54.0 (2.0) \\
& 32 & 56.3 & \textbf{63.5} & 50.2 (1.8) & 47.8 (2.2) & 52.2 (1.2) & 54.9 (1.1) & 53.8 (2.1) \\
\hline
\end{tabular}
}
\caption{Evaluation results across seven vision-language datasets using 0, 4, and 32 in-context examples. 
``OF-3B (I)'' refers to \oftb (Instruct), the 3B model trained with a language-instruction-tuned backbone, while ``Fl-3B'' refers to \fl-3B. \fl results taken from \citet{alayrac2022flamingo}. The highest number in each row is bolded. Full results (including 8- and 16-shot performance) are in Table~\ref{tab:eval_large_norices}.\vspace{-0.1em}}
\label{tab:eval_short_rices}
\end{table*}

\paragraph{Effect of language instruction-tuning.}
We train two \of models at each of the 3B and 4B scales: one model using a base language model, and one with an instruction-tuned variant of the same language model.
In the lower right corner of Figure \ref{fig:results}, we observe that the instruction-tuned variants of MPT-1B and RedPajama-3B on average outperform the base models. The difference is starkest for RedPajama-3B.
Transfer of language instruction tuning to vision-language tasks was previously reported in~\citet{huang2023kosmos, Li2023BLIP2BL}.

\begin{figure}[tb]
    \centering
    \includegraphics[width=\linewidth]{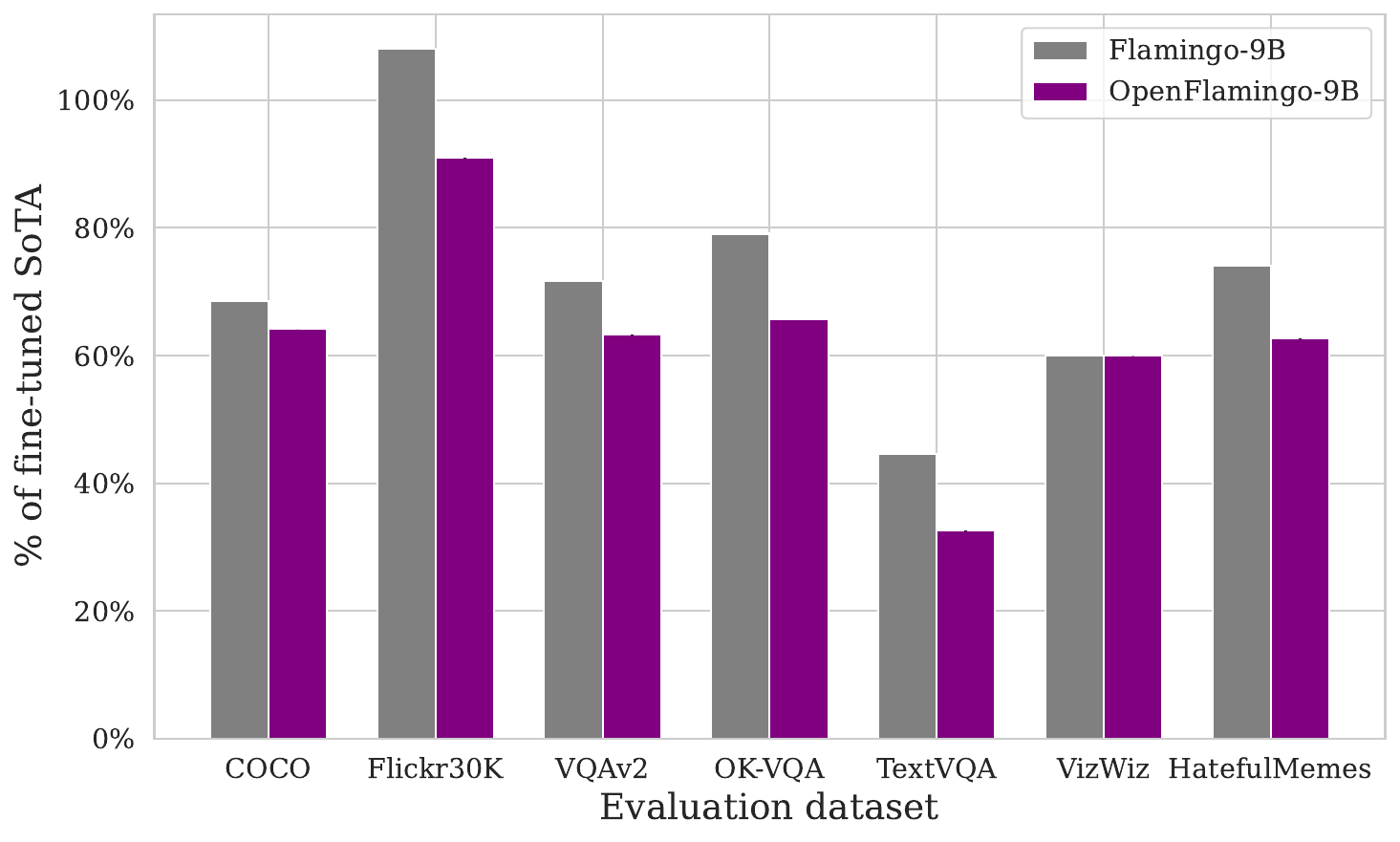}
    \caption{\ofnb and \fl-9B performance relative to fine-tuned SoTA performance.\vspace{-0.5em}}
    \label{fig:relative}
\end{figure}

\paragraph{Comparison to fine-tuned state-of-the-art.} Figure \ref{fig:relative} plots each model's performance relative to fine-tuned state-of-the-art performance, as listed on \href{https://paperswithcode.com/}{Papers With Code} on June 19, 2023. \ofnb averages more than 62\% of fine-tuned state-of-the-art performance with 32 RICES-selected in-context examples, compared to 72\% achieved by \fl-9B.
For more details on the fine-tuned SoTAs, see Appendix \ref{app:sotas}.
\section{Discussion}

\subsection{Frozen embeddings}\label{sec:frozen}
In \S\ref{sec:results}, we observed that \offb models underperform their 3B counterparts on most datasets.
One notable way the \offb models differ from the 3B and 9B models is that their \image and \eoc embeddings are randomly initialized and frozen, rather than trained.

\begin{table}
\centering
\small
\resizebox{\linewidth}{!}{
\begin{tabular}{|c|c|c|c|c|}
\hline
&& \textbf{0-shot} & \textbf{4-shot} & \textbf{8-shot}\\
\hline
\multirow{2}{*}{\textbf{\coco}} & trainable & 46.5 & 58.6 & 61.2 \\
\cline{2-5}
& frozen & 41.9 \gain{41.9}{46.5} & 54.5 \gain{54.5}{58.6} & 57.4 \gain{57.4}{61.2}\\
\hline
\multirow{2}{*}{\textbf{\vqav}} & trainable & 17.6 & 23.2 & 28.7 \\
\cline{2-5}
&frozen& 5.5 \gain{5.5}{17.6} & 8.4 \gain{8.4}{23.2} & 18.8 \gain{18.8}{28.7} \\
\hline
\end{tabular}
}
\caption{\coco and \vqav validation performance when using trainable \image and \eoc embeddings compared to frozen, randomly initialized embeddings. The model used in this experiment is based on CLIP ViT-L/14 and OPT 125M, with cross-attention  every layer, and trained on 20M interleaved samples, including ChatGPT-sequences.}
\label{tab:frozen}
\end{table}

In Table \ref{tab:frozen}, we investigate the effect of this difference.
We train small models using OPT-125M as a language model \cite{Zhang2022OPTOP} to 20M interleaved samples (one-third of full training).
Freezing the \image and \eoc embeddings results in a drop of 4.6 CIDEr for 0-shot \coco, and 12.1\% accuracy for 0-shot \vqav.
This suggests that frozen \image and \eoc embeddings may impact downstream trends.

\subsection{\vqav validation trends}\label{sec:vqa}
During development, we used the \vqav validation set as a temperature check for visual question answering capabilities.
In this section, we discuss trends observed during development.

\begin{table*}[t]
  \centering
  \small
  \setlength{\tabcolsep}{6pt} 
    \renewcommand{\arraystretch}{1} 
    \resizebox{\linewidth}{!}{
  \begin{tabular}{|c|c|c|}
    \multicolumn{1}{c}{ \textbf{Counting}} & \multicolumn{1}{c}{ \textbf{Verbosity}} & \multicolumn{1}{c}{ \textbf{Non-central object}}\\
    \cline{1-3}
    &&\\
    \includegraphics[height=2.75cm]{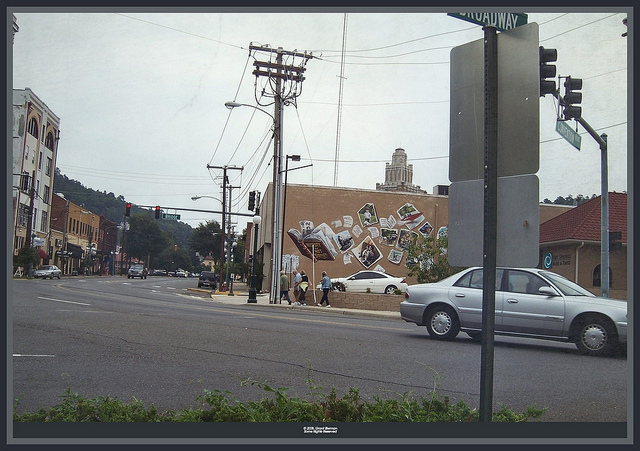} & \includegraphics[height=2.75cm]{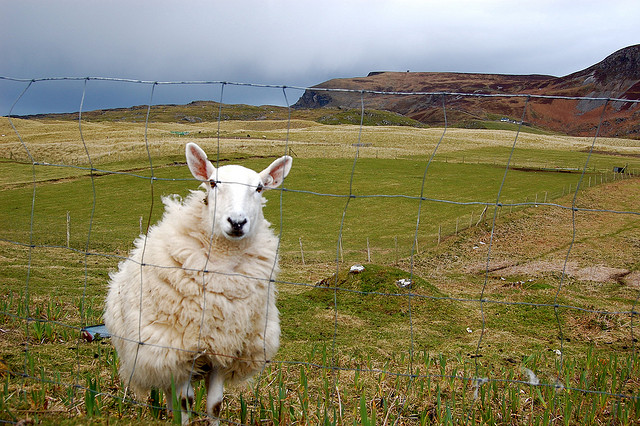}  & \includegraphics[height=2.75cm]{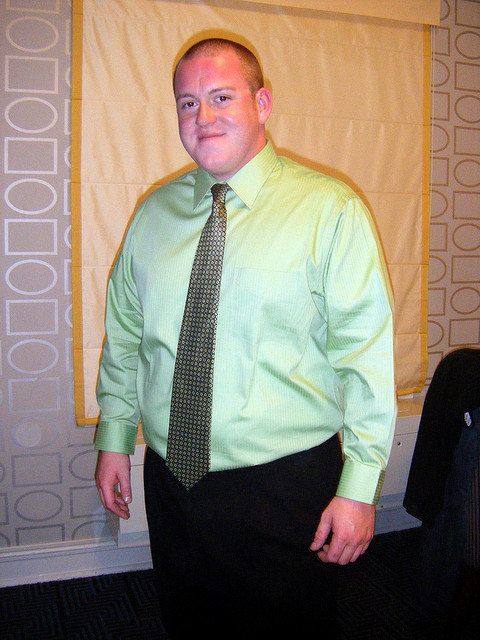}  \\
    {\sc Q:} How many people are on the sidewalk? & {\sc Q:} What is this sheep trying to do? & {\sc Q:} What color are the curtains?\\
    \rule{5cm}{0.2pt}&\rule{5cm}{0.2pt}&\rule{5cm}{0.2pt}\\
    {OF-9B:} ``one'' &  {OF-9B:} ``it is trying to get'' & {OF-9B:} ``green''\\
    \rule{5cm}{0.2pt}&\rule{5cm}{0.2pt}&\rule{5cm}{0.2pt}\\
    {\sc  Ground truth:} \textit{ \{``4'', ``5''\} } & {\sc Ground truth:}\textit{ \{``get out'', ``escape''\} }& {\sc Ground truth:} \textit{\{``yellow'', ``gold''\}}\\
    &&\\
    \cline{1-3}
  \end{tabular}
  }
  \caption{\ofnb errors from the \vqav validation split. Common failure modes for \of including counting, giving answers that are too verbose (and thus truncated), and answering about the central object in the image rather than the non-central object in the question.}
\label{tab:ofnb_vqa_val}
\end{table*}

\begin{figure}
\centering
\includegraphics[width=0.5\textwidth]{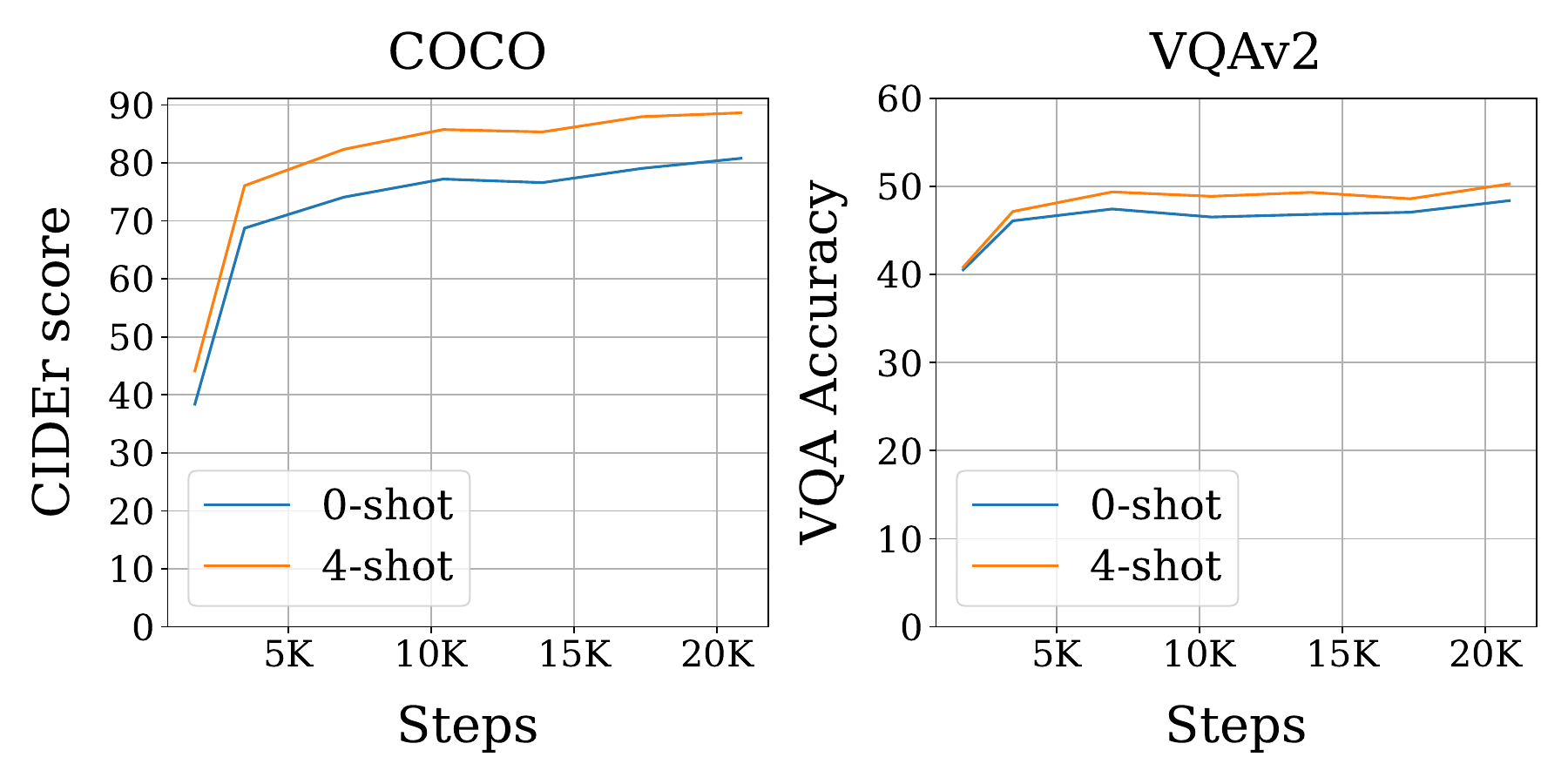}
\caption{Validation split performance for \ofnb across training: while \coco CIDEr improves throughout training, \vqav performance is more stagnant.}
\label{fig:eval_evo}
\end{figure}

\paragraph{Training dynamics.} 
To understand how evaluation performance evolves over the course of training, Figure \ref{fig:eval_evo} plots validation performance of \ofnb on \coco and \vqav throughout training.
While \coco performance steadily improves, \vqav progress is flatter.
This matches trends reported by \citet{Li2023BLIP2BL}.

\paragraph{Effect of language model.}
Although additional training did not dramatically affect \vqav performance, changing language model backbones did.
Table \ref{tab:vqa_lm} illustrates this effect on the \vqav validation split; notably, switching from OPT-1.3B to MPT-1B (Instruct) added nearly 10 percentage points in 0-shot performance.
We hypothesize that the language model has similarly large effects for other VQA tasks.

\begin{table}
\small
\centering
\begin{tabular}{|l|p{1.5cm}|p{1.5cm}|}
\hline
\multirow{3}{*}{} & \multicolumn{2}{c|}{\textbf{\vqav validation}}\\
\cline{2-3}
 & \multicolumn{2}{c|}{\emph{Shots}} \\
\cline{2-3}
\textbf{Language model} & \textbf{0} & \textbf{4} \\
\hline
OPT-125M & 17.6 & 23.2 \\
\hline
OPT-1.3B & 32.8 & 27.2  \\
\hline
MPT-1B (Instruct) & 41.9 & 43.7  \\
\hline
MPT-7B & 47.4 & 49.4  \\
\hline
\end{tabular}
\caption{\vqav validation performance at 20M interleaved samples across different language models. Performance largely differs between language models.\vspace{-0em}}
\label{tab:vqa_lm}
\end{table}

\paragraph{Common VQA failure modes (Table \ref{tab:ofnb_vqa_val}).} 
\of models struggle with counting; on the \vqav validation split, \ofnb scores 30.5\% on questions with numerical answers, compared to 70.6\% on yes / no questions.
Additionally, because VQA accuracy uses an exact match criterion for generations, models must answer concisely to score well; \of models are often too verbose.
Finally, VQA questions can ask about objects other than the central object in the image; models sometimes answer about the central item instead.

\subsection{Applications of \of}
Multiple models have already developed on top of \of. 
\citet{Li2023OtterAM} fine-tuned \of on MIMIC-IT~\cite{Li2023MIMICITMI}, a multi-image/video instruction following dataset, creating Otter, a multimodal assistant. \citet{Gong2023MultiModalGPTAV} released Multimodal-GPT, an \of model instruction fine-tuned on both vision and language instruction datasets. 
We hope the community continues to use \of models.

\subsection{Limitations}
OpenFlamingo models carry the same risks as their foundational language models. In particular, these models train on web-scraped data, and they have not undergone safety-focused fine-tuning. Models thus may produce unexpected, inappropriate, or inaccurate outputs. 
We hope to further investigate the safety properties of \typemodels like \of.
\section{Conclusion}
In this technical report, we described \of, a family of five \typemodels across the 3B, 4B, and 9B scales.
\of remains an active research project, and we continue to work on training and releasing high-quality \typemodels. 
We hope our contribution enables more researchers to train and study such models.


{\small
\subsection*{Acknowledgements}
We would like to thank Jean-Baptiste Alayrac and Antoine Miech for their advice on reproducing Flamingo. We also thank Rohan Taori, Nicholas Schiefer, Deep Ganguli, Thomas Liao, Tatsunori Hashimoto, and Nicholas Carlini for their help with assessing the safety risks of our first release of \of.
Thanks to Stability AI for compute resources.
{\small
\bibliographystyle{plainnat}
\bibliography{main}
}
\clearpage
\appendix

\begin{table}[tbh]
\small
\centering
\caption{Fine-tuned state-of-the-art numbers used in this report. \label{tab:sotas}}
\begin{tabular}{@{}lll@{}}
\toprule
\multicolumn{1}{l}{\textbf{Method}} & \textbf{Dataset} & \textbf{Score} \\ \midrule
                                 mPLUG~\cite{Li2022mPLUGEA}   & \coco             &   155.1             \\
                                Unified VLP~\cite{Zhou2019UnifiedVP}    & \flickr        &    67.4          \\
                                Pali-17B~\cite{chen2022pali}    & \vqav            &       84.3         \\
                                Pali-17B~\cite{chen2022pali}    & \okvqa           &   64.5             \\
                                Pali-17B~\cite{chen2022pali}    & \textvqa          &     73.1           \\
                                Pali-17B~\cite{chen2022pali}    & \vizwiz           &  73.3              \\
                                 Hate-CLIPper~\cite{Kumar2022HateCLIPperMH}   & \hms     &   85.8             \\ \bottomrule
\end{tabular}%
\end{table}

\begin{table}[tb]
\centering
\resizebox{\linewidth}{!}{
\begin{tabular}{l|c||r|r}
\textbf{Benchmark} & \textbf{Shots} & \textbf{Random} & \textbf{RICES} \\ 
\hline
\multirow{2}{*}{\textbf{\coco}} 
& 4  & 89.0 & 93.1 \gain{93.1}{89}\\
& 32 & 99.5 & 109.0 \gain{109.0}{99.5} \\
\hline
\multirow{2}{*}{\textbf{\flickr}} 
& 0 & 59.5& 39.2 \gain{39.2}{59.5}\\
& 4 & 65.8 & 52.2 \gain{52.2}{65.8}\\
& 8 & 62.9 & 58.7 \gain{58.7}{62.9} \\
& 32 & 61.3 & 63.0 \gain{63.0}{61.3}\\
\hline
\multirow{2}{*}{\textbf{\vqav}} 
& 4 &  54.8 & 55.1 \gain{55.1}{54.8}\\
& 32 & 53.3 & 56.8 \gain{56.8}{53.3}\\
\hline
\multirow{2}{*}{\textbf{\okvqa}} 
& 4 & 40.1 & 38.3 \gain{38.3}{40.1}\\
& 32 & 42.4 & 46.3 \gain{46.3}{42.4}\\
\hline
\multirow{2}{*}{\textbf{\textvqa}} 
& 4  & 28.2 & 34.2 \gain{34.2}{28.2}\\
& 32 & 23.8 & 31.1 \gain{31.1}{23.8}\\
\hline
\multirow{2}{*}{\textbf{\vizwiz}} 
& 4 & 27.5 & 41.0 \gain{41.0}{27.5}\\
& 32 & 44.0 & 46.4 \gain{46.4}{44.0}\\
\hline
\multirow{2}{*}{\textbf{\hms}} 
& 4 & 54.0 & 70.1 \gain{70.1}{54.0}\\
& 32 & 53.8 & 73.6 \gain{73.6}{53.8}\\
\hline
\end{tabular}
}
\caption{Using RICES \cite{yang2022empirical} to select in-context examples often outperforms using random demonstrations. Scores in table are for \ofnb.}
\label{tab:rices_v_random}
\end{table}

\begin{table*}[ht]
\centering
\small
\setlength{\tabcolsep}{6pt} 
\renewcommand{\arraystretch}{1.0} 
\begin{tabular}{r|p{6cm}|p{6cm}|}
\cline{2-3} & \multicolumn{1}{c|}{\textbf{Random demonstrations}}&  \multicolumn{1}{c|}{\textbf{RICES}}\\
\cline{2-3}
\multirow{8}{*}{Demos} & \multicolumn{1}{c|}{\includegraphics[height=1.5cm]{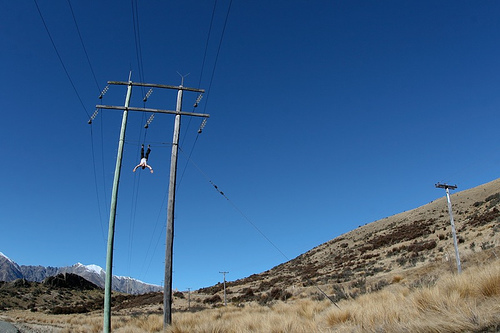}} & \multicolumn{1}{c|}{\includegraphics[height=1.5cm]{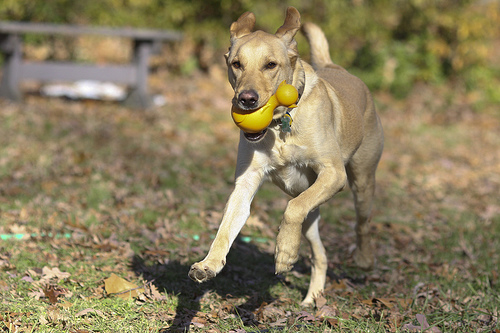}} \\
& \scriptsize A person hanging from a telephone pole near the mountains. & \scriptsize  The brown dog is running through the grass with a yellow toy in its mouth. \\
& \multicolumn{1}{c|}{\includegraphics[height=1.5cm]{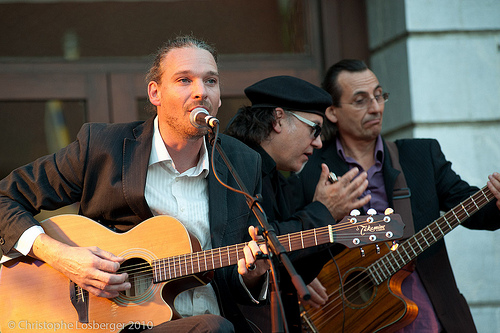}} & \multicolumn{1}{c|}{\includegraphics[height=1.5cm]{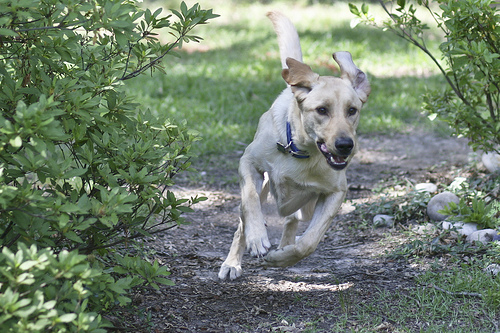}} \\
& \scriptsize A trio of male musicians are performing with one playing a guitar and singing into a microphone, another holding a harmonica, and the third playing a bass guitar. & \scriptsize  A white dog rushes down a dirt path surrounded by grass and trees. \\
& \multicolumn{1}{c|}{\includegraphics[height=1.5cm]{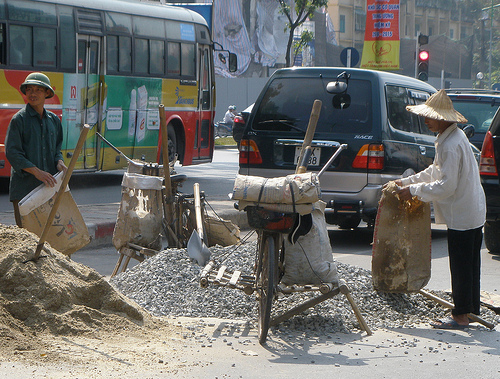}} & \multicolumn{1}{c|}{\includegraphics[height=1.5cm]{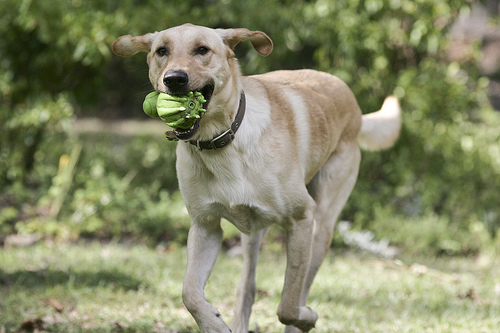}} \\
& \scriptsize Two men, both in strange hats, working over rocks in a busy urban street. & \scriptsize The tan dog is carrying a green squeak toy in its mouth.\\
& \multicolumn{1}{c|}{\includegraphics[height=1.5cm]{figures/flickr/random_4.png}} & \multicolumn{1}{c|}{\includegraphics[height=1.5cm]{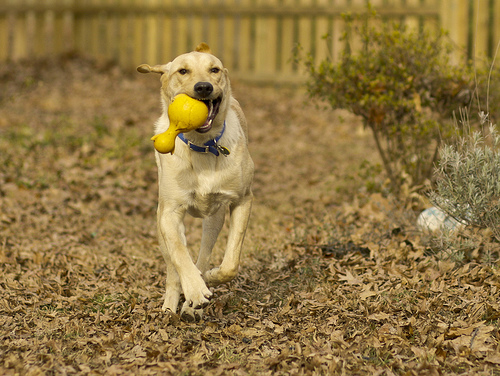}} \\
& \scriptsize Several people are in a group where a man in a blue shirt is smiling. & \scriptsize A yellow dog running through a yard covered in leaves while holding a yellow toy in his mouth. \\
& \multicolumn{1}{c|}{\rule{5cm}{0.2pt}} & \multicolumn{1}{c|}{\rule{5cm}{0.2pt}} \\
Test example & \multicolumn{1}{c|}{\includegraphics[height=2.5cm]{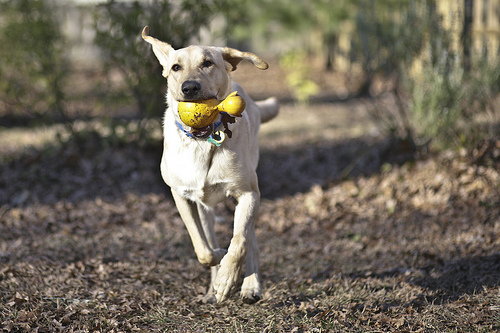}} & \multicolumn{1}{c|}{\includegraphics[height=2.5cm]{figures/flickr/test.png}} \\
{\scriptsize OF-9B {generations:}} & \footnotesize {A yellow labrador retriever running with a ball.} & \footnotesize {A yellow dog running through a yard covered in leaves while holding a green squeak toy in his mouth} \\
\cline{2-3}
& \multicolumn{2}{c|}{\scriptsize \emph{{\sc Ground truth:} A white dog fetching a yellow toy.} }   \\
\cline{2-3}
\end{tabular}
\caption{Comparison of \ofnb outputs for a \flickr 4-shot evaluation using RICES vs. random demonstrations. With RICES, \ofnb patches together these demonstration captions to answer for the test image, including incorrect details.}
\label{tab:ofnb_flickr_rices}
\end{table*}

\section{Extended results}\label{app:more_eval}
Table \ref{tab:eval_large_norices} provides full evaluation results for 0, 4, 8, 16, and 32 in-context examples.
For ease of comparison to \fl, we calculate each \of model's performance as a fraction of corresponding \fl performance in Figure \ref{fig:percentage}.

\subsection{Comparison to fine-tuned SoTAs}\label{app:sotas}
In Figure \ref{fig:agg}, we compare \of models to fine-tuned SoTA performances for different numbers of in-context examples. 
The fine-tuned methods used were pulled from PapersWithCode on 06/19/23 (Table \ref{tab:sotas}).

\begin{figure}[h]
\centering
\includegraphics[width=\linewidth]{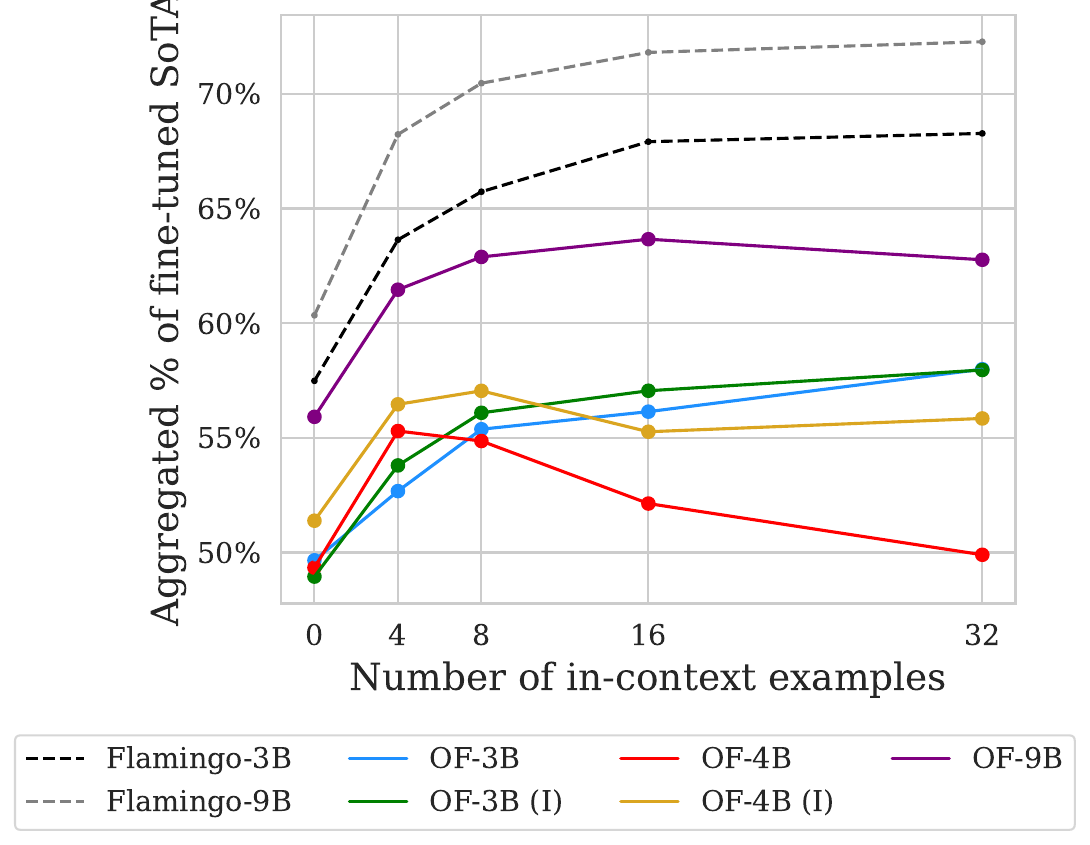}
\caption{We plot each model's performance relative to fine-tuned state-of-the-art performance, averaged across datasets.\vspace{-1em}}
\label{fig:agg}
\end{figure}

\subsection{Evaluations using RICES}\label{app:rices}
In the main text, we evaluate \of by selecting in-context examples uniformly at random.
In this appendix, we include additional evaluation results using Retrieval-based In-Context Example Selection (RICES) \cite{yang2022empirical}.
For a given test example, RICES selects the top-k most similar training examples as demonstrations, where similarity is measured by cosine similarity of the images according to the frozen vision encoder (CLIP ViT-L/14).
Full results with RICES are listed in Table \ref{tab:eval_large_rices} and illustrated in Figure \ref{fig:results_rices}.

In Table \ref{tab:rices_v_random}, we compare \ofnb performance using RICES to performance using randomly selected in-context examples.
We observe that RICES significantly boosts performance in most evaluation settings, including by 19.2 ROC AUC using 32 shots on \hms.
However, on \flickr, we observe significant degradations from using RICES: 
CIDEr degrades by 20.4 in 0-shot evaluations\footnote{In 0-shot evaluations, RICES is still used to select the two text-only examples used for the prompt (\S\ref{sec:eval_method}).} and 13.1 in 4-shot evaluations.
We hypothesize that the demonstrations RICES selects in \flickr are more similar to the test example than in other datasets. 
This leads \ofnb to parrot captions from the in-context examples, including incorrect details. 
For an example, see Table \ref{tab:ofnb_flickr_rices} in Appendix \ref{app:more_eval}.

\begin{figure*}[bt] 
  \centering
  \includegraphics[width=\linewidth]{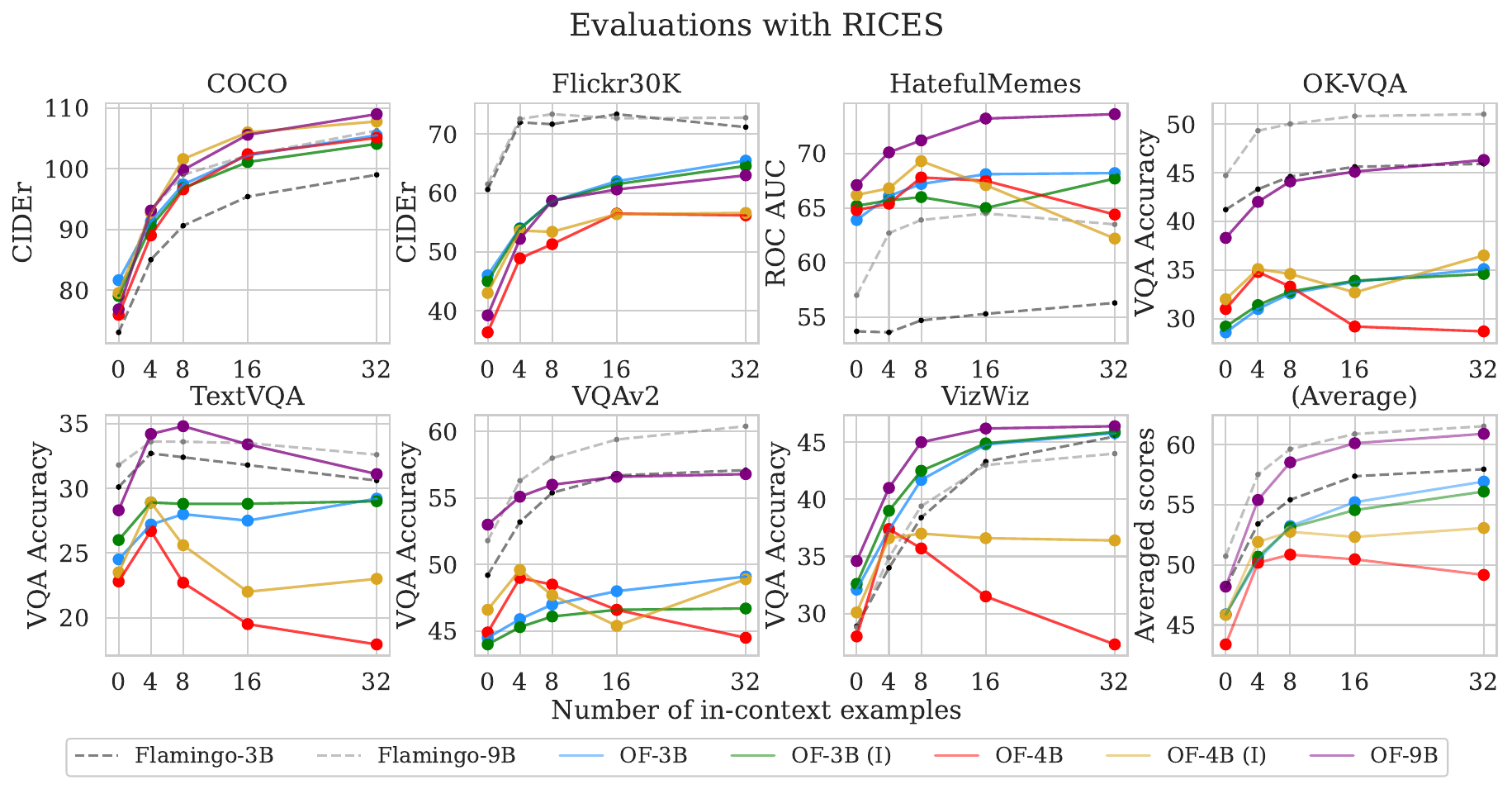} 
  \caption{Evaluation results per dataset across 0, 4, 8, 16, and 32 in-context examples. Results are reported in tabular form in Table \ref{tab:eval_large_rices}.\vspace{-0.1em}}
  \label{fig:results_rices}
\end{figure*}

\begin{figure*}[t]
    \centering
    \includegraphics[width=0.8\linewidth]{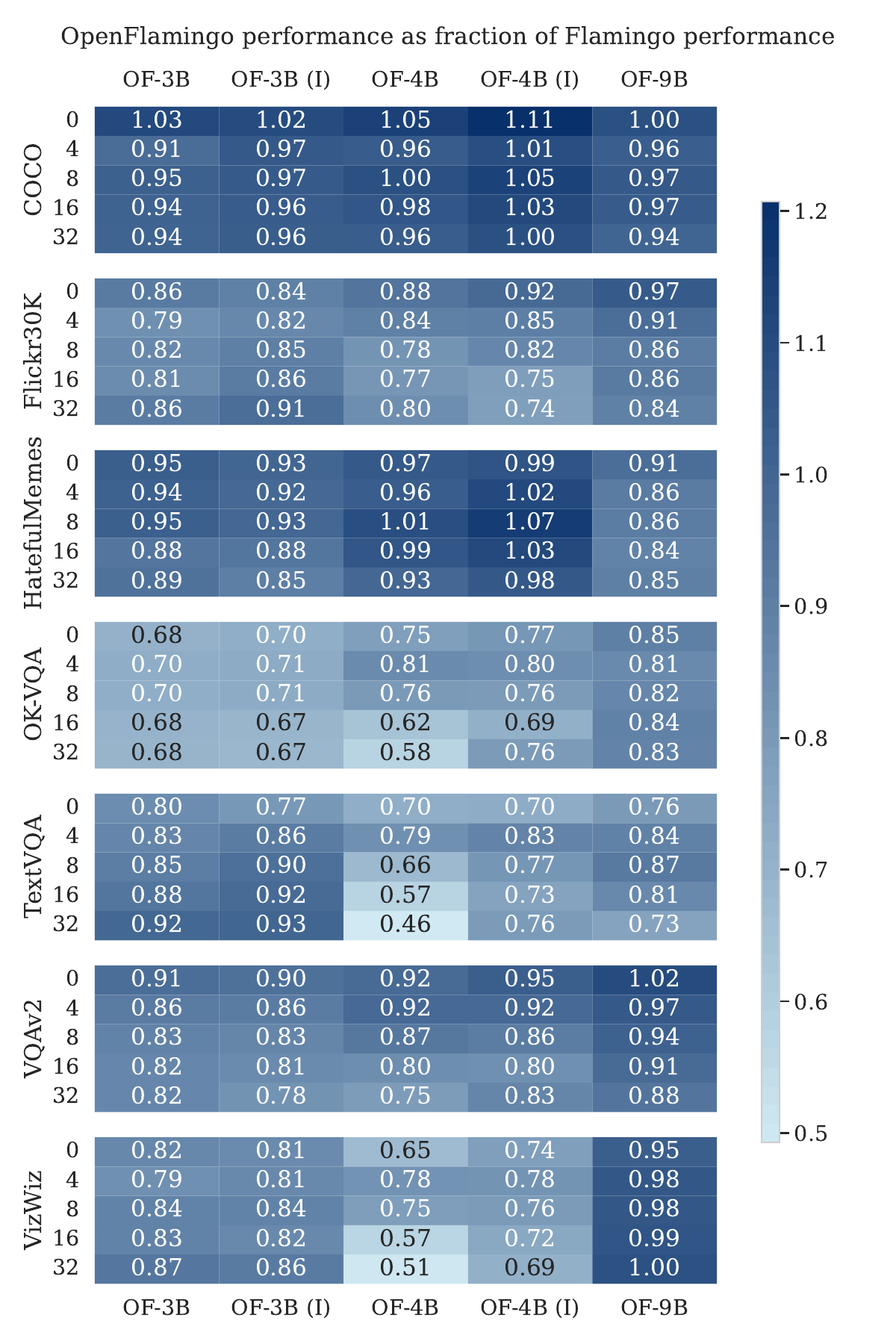}
    \caption{\of performance as a fraction of corresponding \fl performance for each evaluation setting. We compare \of-3B and -4B models to \fl-3B, and \ofnb to \fl-9B. 0, 4, 8, 16, and 32 refer to the number of in-context examples used.}
    \label{fig:percentage}
\end{figure*}

\begin{table*}[ht]
\centering
\resizebox{\textwidth}{!}{
\begin{tabular}{l|c||c|c||c|c|c|c|c}
\textbf{Benchmark} & \textbf{Shots} & \textbf{Fl-3B} & \textbf{Fl-9B} & \textbf{OF-3B} & \textbf{OF-3B (I)} & \textbf{OF-4B} & \textbf{OF-4B (I)} & \textbf{OF-9B}\\ 
\hline
\multirow{5}{*}{\textbf{\coco} \cite{Chen2015MicrosoftCC}} & 0  & 73.0 & 79.4 & 74.9 (0.2) & 74.4 (0.6) & 76.7 (0.2) & \textbf{81.2 (0.3)} & 79.5 (0.2)\\
& 4  & 85.0 & \textbf{93.1} & 77.3 (0.3) & 82.7 (0.7) & 81.8 (0.4) & 85.8 (0.5) & 89.0 (0.3)\\
& 8 & 90.6 & \textbf{99.0} & 85.9 (0.6) & 87.8 (0.5) & 90.7 (0.3) & 94.8 (0.2) & 96.3 (0.1)\\
& 16 & 95.4 & \textbf{102.2} &  89.8 (0.2) & 91.9 (0.3) & 93.9 (0.4)  & 98.0 (0.3) & 98.8 (0.7) \\
& 32 & 99.0 & \textbf{106.3} &  93.0 (0.6) & 94.8 (0.3) & 95.1 (0.3) & 99.2 (0.3) & 99.5 (0.1)\\
\hline
\multirow{5}{*}{\textbf{\flickr} \cite{Young2014FromID}} & 0 & 60.6 & \textbf{61.5} & 52.3 (1.0) & 51.2 (0.2) & 53.6 (0.9) & 55.6 (1.3) & 59.5 (1.0)\\
& 4 & 72.0 & \textbf{72.6} & 57.2 (0.4) & 59.1 (0.3) & 60.7 (1.2) & 61.2 (0.5) & 65.8 (0.6)\\
& 8 & 71.7 & \textbf{73.4} &  58.6 (1.1) & 60.7 (0.6) & 55.9 (1.3) & 59.0 (1.0) & 62.9 (1.0)\\
& 16 & \textbf{73.4} & 72.7 & 59.2 (0.5) & 63.0 (0.4) & 56.8 (0.5) & 54.8 (1.0) & 62.8 (1.0)\\
& 32 & 71.2 & \textbf{72.8} &  61.1 (1.3) & 64.5 (1.3) & 56.9 (0.7) & 53.0 (0.5) & 61.3 (0.7)\\
\hline
\multirow{5}{*}{\textbf{\vqav} \cite{Agrawal2015VQAVQ}} 
& 0 & 49.2 & 51.8 &  44.6 (0.0) & 44.1 (0.1) & 45.1 (0.1) & 46.9 (0.0) & \textbf{52.7 (0.2)}\\
& 4 & 53.2 & \textbf{56.3} & 45.8 (0.0) & 45.7 (0.1) & 49.0 (0.0) & 49.0 (0.0) & 54.8 (0.0)\\
& 8 & 55.4 & \textbf{58.0} & 46.2 (0.0) & 45.9 (0.1) & 48.3 (0.0) & 47.4 (0.0) & 54.8 (0.0)\\
& 16 & 56.7 & \textbf{59.4} & 46.6 (0.0) & 45.8 (0.0) & 45.5 (0.1) & 45.1 (0.1) & 54.3 (0.0) \\
& 32 & 57.1 & \textbf{60.4} & 47.0 (0.1) & 44.8 (0.1) & 43.0 (0.2) & 47.3 (0.0) & 53.3 (0.1)\\
\hline
\multirow{5}{*}{\textbf{\okvqa} \cite{Marino2019OKVQAAV}} 
& 0 & 41.2 & \textbf{44.7} & 28.2 (0.2) & 28.7 (0.1) & 30.7 (0.1) & 31.7 (0.1) & 37.8 (0.2)\\
& 4 & 43.3 & \textbf{49.3} & 30.3 (0.5) & 30.6 (0.2) & 35.1 (0.0) & 34.6 (0.0) & 40.1 (0.1)\\
& 8 & 44.6 & \textbf{50.0} & 31.1 (0.3) & 31.5 (0.3) & 33.9 (0.1) & 33.7 (0.2) & 41.1 (0.2) \\
& 16 & 45.6 & \textbf{50.8} & 30.9 (0.3) & 30.7 (0.3)  & 28.5 (0.2) & 31.3 (0.1) & 42.7 (0.2) \\
& 32  & 45.9 & \textbf{51.0} & 31.0 (0.1) & 30.6 (0.1) & 26.4 (0.2) & 34.7 (0.3) & 42.4 (0.0)\\
\hline
\multirow{5}{*}{\textbf{\textvqa} \cite{Singh2019TowardsVM}} & 0 & 30.1 & \textbf{31.8} & 24.2 (0.2) & 23.1 (0.2) & 21.0 (0.3) & 21.1 (0.4) & 24.2 (0.5)\\
& 4  & 32.7 & \textbf{33.6} &  27.0 (0.3) & 28.1 (0.4) & 25.9 (0.0) & 27.2 (0.3)  & 28.2 (0.4)\\
& 8 & 32.4 & \textbf{33.6} & 27.7 (0.1) & 29.1 (0.1) & 21.3 (0.2) & 25.1 (0.2) & 29.1 (0.1)\\
& 16 & 31.8 & \textbf{33.5} & 28.0 (0.2)  & 29.1 (0.1) & 18.2 (0.4) & 23.2 (0.1) & 27.3 (0.1)\\
& 32 & 30.6 & \textbf{32.6} & 28.3 (0.2) & 28.5 (0.1) & 14.1 (0.2) & 23.2 (0.2) & 23.8 (0.2)\\
\hline
\multirow{5}{*}{\textbf{\vizwiz} \cite{Gurari2018VizWizGC}} 
& 0 & \textbf{28.9} & 28.8 & 23.7 (0.5)  & 23.4 (0.3) & 18.8 (0.1) & 21.5 (0.2) & 27.5 (0.2)\\
& 4 & 34.0 & \textbf{34.9} &27.0 (0.3) & 27.7 (0.1) & 26.6  (0.5)& 26.5 (0.4) & 34.1 (0.7)\\
& 8 & 38.4 & \textbf{39.4} & 32.1 (0.7) & 32.1 (0.6) & 28.8 (0.4)& 29.1 (0.2) & 38.5 (0.1)\\
& 16 & \textbf{43.3} & 43.0 & 36.1 (0.3) & 35.3 (0.1) & 24.6 (0.2) & 31.0 (0.6) & 42.5 (0.4) \\
& 32 & \textbf{45.5} & 44.0 & 39.8 (0.1) & 39.3 (0.4) & 23.1 (1.1) & 31.3 (0.2) & 44.0 (0.5)\\
\hline
\multirow{5}{*}{\textbf{\hms} \cite{Kiela2020TheHM}} & 0 & 53.7 & \textbf{57.0} & 51.2 (2.5) & 50.1 (2.2) & 52.3 (2.3) & 53.1 (2.2) & 51.6 (1.8) \\
& 4 & 53.6 & \textbf{62.7} & 50.6 (0.8) & 49.5 (0.6) & 51.5 (1.4) & 54.9 (1.1) & 54.0 (2.0) \\
& 8 & 54.7 & \textbf{63.9} & 52.0 (1.1) & 50.7 (1.8) & 55.2 (0.8) & 58.5 (0.3) & 54.7 (2.8)\\
& 16 & 55.3 & \textbf{64.5} & 48.5 (0.7)  & 48.7 (1.0) & 54.5 (1.3) & 56.9 (1.5) & 53.9 (3.1)\\
& 32 & 56.3 & \textbf{63.5} & 50.2 (1.8) & 47.8 (2.2) & 52.2 (1.2) & 54.9 (1.1) & 53.8 (2.1) \\
\hline
\end{tabular}
}
\caption{Full evaluation results using \textbf{demonstrations sampled uniformly at random} across seven vision-language datasets using 0, 4, 8, 16, and 32 in-context examples. Results are averaged across 3 evaluation seeds and reported with standard deviations.}
\label{tab:eval_large_norices}
\end{table*}
\begin{table*}[ht]
\centering
\resizebox{\textwidth}{!}{
\begin{tabular}{l|c||c|c||c|c|c|c|c}
\textbf{Benchmark} & \textbf{Shots} & \textbf{Fl-3B} & \textbf{Fl-9B} & \textbf{OF-3B} & \textbf{OF-3B (I)} & \textbf{OF-4B} & \textbf{OF-4B (I)} & \textbf{OF-9B}\\ 
\hline
\multirow{5}{*}{\textbf{\coco} \cite{Chen2015MicrosoftCC}} & 0 & 73.0 & 79.4 & \textbf{81.6} & 79.0 & 75.9 & 79.5 & 76.8 \\
& 4 & 85.0 & \textbf{93.1} & 91.3 & 90.5 & 89.0 & 92.7 & \textbf{93.1} \\
& 8 & 90.6 & 99.0 & 97.4 & 96.8 & 96.6 & \textbf{101.6} & 99.8 \\
& 16 & 95.4 & 102.2 & 102.2 & 101.1 & 102.4 & \textbf{106.0} & 105.6\\
& 32 & 99.0 & 106.3 & 105.5 & 104.1 & 105.1 & 107.8 & \textbf{109.0} \\
\hline
\multirow{5}{*}{\textbf{\flickr} \cite{Young2014FromID}} & 0 & 60.6 & \textbf{61.5} & 46.0 & 45.0 & 36.3 & 43.0 & 39.2 \\
& 4 & 72.0 & \textbf{72.6} & 54.0 & 53.9 & 48.9 & 53.6 & 52.2 \\
& 8 & 71.7 & \textbf{73.4} & 58.6 & 58.6 & 51.3 & 53.4 & 58.7 \\
& 16 & \textbf{73.4} & 72.7 & 62.0 & 61.5 & 56.5 & 56.4 & 60.6 \\
& 32 & 71.2 & \textbf{72.8} & 65.5 & 64.6 & 56.2 & 56.6 & 63.0 \\
\hline
\multirow{5}{*}{\textbf{\vqav} \cite{Agrawal2015VQAVQ}} & 0 & 49.2 & 51.8 & 44.5 & 44.0 & 44.9 & 46.6 & \textbf{53.0} \\
& 4 & 53.2 & \textbf{56.3} & 45.9 & 45.3 & 49.0 & 49.6 & 55.1 \\
& 8 & 55.4 & \textbf{58.0} & 47.0 & 46.1 & 48.5 & 47.7 & 56.0 \\
& 16 & 56.7 & \textbf{59.4} & 48.0 & 46.6 & 46.6 & 45.4 & 56.6 \\
& 32 & 57.1 & \textbf{60.4} & 49.1 & 46.7 & 44.5 & 48.9 & 56.8 \\
\hline
\multirow{5}{*}{\textbf{\okvqa} \cite{Marino2019OKVQAAV}} & 0 & 41.2 & \textbf{44.7} & 28.6 & 29.2 & 31.0 & 32.0 & 38.3 \\
& 4 & 43.3 & \textbf{49.3} & 31.0 & 31.4 & 34.8 & 35.1 & 42.0 \\
& 8 & 44.6 & \textbf{50.0} & 32.6 & 32.8 & 33.3 & 34.6 & 44.1 \\
& 16 & 45.6 & \textbf{50.8} & 33.8 & 33.9 & 29.2 & 32.7 & 45.1 \\
& 32 & 45.9 & \textbf{51.0} & 35.1 & 34.6 & 28.7 & 36.5 & 46.3 \\
\hline
\multirow{5}{*}{\textbf{\textvqa} \cite{Singh2019TowardsVM}} & 0 & 30.1 & \textbf{31.8} & 24.5 & 26.0 & 22.8 & 23.5 & 28.3 \\
& 4 & 32.7 & 33.6 & 27.2 & 28.9 & 26.7 & 28.9 & \textbf{34.2} \\
& 8 & 32.4 & 33.6 & 28.0 & 28.8 & 22.7 & 25.6 & \textbf{34.8} \\
& 16 & 31.8 & \textbf{33.5} & 27.5 & 28.8 & 19.5 & 22.0 & 33.4 \\
& 32 & 30.6 & \textbf{32.6} & 29.2 & 29.0 & 17.93 & 23.0 & 31.1 \\
\hline
\multirow{5}{*}{\textbf{\vizwiz} \cite{Gurari2018VizWizGC}} & 0 & 28.9 & 28.8 & 32.1 & 32.6 & 28.0 & 30.1 & \textbf{34.6} \\
& 4 & 34.0 & 34.9 & 37.3 & 39.0 & 37.4 & 36.6 & \textbf{41.0} \\
& 8 & 38.4 & 39.4 & 41.7 & 42.5 & 35.7 & 37.0 & \textbf{45.0} \\
& 16 & 43.3 & 43.0 & 44.8 & 44.9 & 31.5 & 36.6 & \textbf{46.2} \\
& 32 & 45.5 & 44.0 & 45.8 & 45.9 & 27.3 & 36.4 & \textbf{46.4} \\
\hline
\multirow{5}{*}{\textbf{\hms} \cite{Kiela2020TheHM}} & 0 & 53.7 & 57.0 & 63.9 & 65.2 & 64.8 & 66.2 & \textbf{67.1}\\
& 4 & 53.6 & 62.7 & 66.1 & 65.7 & 65.4 & 66.8 & \textbf{70.1}\\
& 8 & 54.7 & 63.9 & 67.2 & 66.0 & 67.8 & 69.3 & \textbf{71.2} \\
& 16 & 55.3 & 64.5 & 68.1 & 65.0 & 67.5 & 67.1 & \textbf{73.2} \\
& 32 & 56.3 & 63.5 & 68.2 & 67.7 & 64.4 & 62.2 & \textbf{73.6} \\
\hline
\end{tabular}
}
\caption{Full evaluation results using RICES across seven vision-language datasets using 0, 4, 8, 16, and 32 in-context examples.}
\label{tab:eval_large_rices}
\end{table*}

\section{Additional notes on filtering MMC4}\label{app:mmc4}
When training contrastive vision-language models, filtering image-text pairs by CLIP cosine similarity has proven particularly helpful for improving data quality  \cite{santurkar2022caption,gadre2023datacomp}.
We use a similar notion for filtering interleaved sequences in MMC4:
if an image and its matched sentence had cosine similarities that fell below a fixed threshold (0.24), according to CLIP ViT-L/14 embeddings, we omitted the image from the sequence, keeping the text.
If all images in a sequence are omitted, we discard the sequence entirely.
This aims to ensure that images are relevant to the text following it. 

However, increasing the image-text similarity threshold has a side effect: it reduces the typical number of images per interleaved sequence.
When using similarity 0.32, nearly 58\% of a sample of 1,000 MMC4 sequences contain only 1 image per sequence, compared to 38\% in Figure \ref{fig:mmc4_hist}, which uses a threshold of 0.24.
Training with long sequences may be important for producing models that can handle a large amount of in-context examples.
Further, we estimate that 88.7\% of MMC4 sequences are discarded completely when filtering with threshold 0.32,
compared to 42.7\% with threshold 0.24.

As future work, we are interested in understanding how to balance length, quality, and dataset size objectives to improve \of models.

\section{Synthetic data prompt}
\label{sec:sec_with_synthetic_prompt}

We provide the prompt used to generate the ChatGPT-generated data (see \S\ref{sec:sec_with_synthetic_data}) in Table~\ref{tab:full_prompt_example_chatgpt}. After generating candidate sequences, we query LAION-5B using \cite{beaumont-2022-clip-retrieval} to infill images. For each unique caption we generate, we attempt to retrieve 10 candidate images from the index using index=laion5B-L-14, aesthetic\_score=9, and aesthetic\_weight=0.5. After this search, we re-rank the retrieved images using CLIP ViT-L/16@336px and select the image with the highest similarity to interleave.

\begin{figure*}
\centering
\resizebox{.95\linewidth}{!}{

\begin{tabular}{p{1.3\linewidth}}

\toprule
\underline{\texttt{system prompt} (human authored)}\\
\vspace{.01cm}
You are a helpful assistant that knows a lot about image descriptions and how they are used with text in broader contexts.\\\\

Your job is to generate an interleaved sequence of images (referenced by their image descriptions) and sentences.\\\\

Image descriptions should only describe the single image. They should be specific, 1-sentence, self-contained descriptions that capture the salient elements of the image.\\\\

The sentences can refer to concepts in any of the images and to each-other. The interleaved sequences are interesting, clever, and represent compelling patterns. To understand them requires knowledge of real-world entities/events, comparison between the images, and commonsense knowledge. Real-world references/facts must be correct.\\\\

Remember to:\\
1) be creative. tell stories, reference real-world entities/events, etc. The images/sentence can play off each-other in fun ways.\\
2) be interesting. generate sequences that are cool, fun, compelling and require interesting commonsense reasoning across and between images/sentences.\\
3) make sure the image descriptions are self-contained, and the output format follows the requested format.\\
\midrule
\underline{\texttt{user} (human authored)}\\
\vspace{.01cm}
Generate a creative, interesting sequence of sentences/images with the following format: (image A, sentence 1, image B, sentence 2, image C, sentence 3) \\

\underline{\texttt{assistant} (human authored)}\\
\vspace{.01cm}
Sure! \\\\
Sequence format: (image A, sentence 1, image B, sentence 2, image C, sentence 3). This interleaved sequence of images/sentences contains images of animals from various parts of the world. The sentences first name the depicted animal, and then describe where the animal is native to.\\\\

Image A: A photo of a person holding a chinchilla in their hand\\
Sentence 1: This is a chinchilla. They are mainly found in Chile.\\
Image B: A photo of a shiba-enu in the fall sitting in the forest\\
Sentence 2: This is a shiba. They are very popular in japan.\\
Image C: A photo of a flamingo with a lake in the background\\
Sentence 3: This is a flamingo. They are found in the Caribbean and South America. \\

\vspace{.01cm}
\texttt{(... 4x more hand designed/interleaved examples...)}\\
\midrule
\vspace{.01cm}

\underline{\texttt{user} (human authored)}\\
\vspace{.01cm}
Generate a creative, interesting sequence of sentences/images with the following format: (image A, image B, sentence 1, image C, image D, sentence 2, image E, image F, sentence 3).\\

\underline{\texttt{assistant} (machine generated)}\\
\vspace{.01cm}
Sure!\\\\

This interleaved sequence of images/sentences contains images of nature. The sentences that follow each image are statements that lead to conclusions. The final sentence explains the broader significance of the pattern.\\
Image A: A photo of a seed being planted in the soil.\\
Image B: A photo of a small plant sprouting from the soil.\\
Sentence 1: The plant is growing.\\
Image C: A photo of trees in a forest.\\
Image D: A photo of an orchard with rows of apple trees.\\
Sentence 2: Trees are used to produce fruits.\\
Image E: A photo of a river running through a valley.\\
Image F: A photo of a waterfall cascading down a mountain.\\
Sentence 3: Water is a critical component of many natural processes, from plant growth to the formation of landscapes.\\

\bottomrule

\end{tabular} }

\caption{We prompt ChatGPT using a system prompt, 5 interleaved human-authored sequences, and a query pattern, in this case ``(Image A, Image B, Sentence 1, Image C, Image D, Sentence 2, Image E, Image F, Sentence 3)". Based on these examples, ChatGPT first explains what sequence it is attempting to generate (here: ``... images of nature. The sentences that follow each image are statements that lead to conclusions"). Later, images are retrieved based on the generated ``Photo X" descriptions to form interleaved sequences.}
\label{tab:full_prompt_example_chatgpt}

\end{figure*}

\begin{table*}[t]
\centering
\begin{tabular}{|c|c|c|}
\hline
\multicolumn{3}{|c|}{Row 1} \\ \hline
\href{http://images.cocodataset.org/val2017/000000039769.jpg}{Link} &
\href{https://twitter.com/AndrewMayne/status/1511827454536474626?s=20}{Link} &
\href{https://cdn.openai.com/multimodal-neurons/assets/apple/apple-ipod.jpg}{Link} \\ \hline
\multicolumn{3}{|c|}{Row 2} \\ \hline
\href{https://encrypted-tbn0.gstatic.com/images?q=tbn:ANd9GcSryNSXyspcZaNkgHbLaST6r2oYiondM0SWWy7-a4GU&s}{Link} &
\href{https://media.istockphoto.com/id/174615872/photo/famous-places-pike-place-market-street-sign.jpg?s=612x612&w=0&k=20&c=I16FJ_-5jHfmNOEWU0xhQPFzKSqez5HGolKhDiC3ouE=}{Link} &
\href{https://encrypted-tbn0.gstatic.com/images?q=tbn:ANd9GcSNhd1MAg_HzjNzhTOovyzEeDe1yilpfTz7_g&usqp=CAU}{Link} \\ \hline
\multicolumn{3}{|c|}{Row 3} \\ \hline
\href{https://www.lattiz.com/sites/default/files/general/2019-08/Inspiration_Howto_s_latteart_theswan_header.jpg}{Link} &
\href{https://static.wikia.nocookie.net/among-us-wiki/images/3/31/Red.png/revision/latest/thumbnail/width/360/height/360?cb=20230601155504}{Link} &
\href{https://cdn.motor1.com/images/mgl/W8n02j/s3/tesla-model-3.jpg}{Link} \\ \hline
\end{tabular}
\caption{Source of images for Figure~\ref{fig:samples} for each row the links are from left to right in the order of how the images are presented.
\label{tab:image_credits_fig_2}}
\end{table*}

\section{Image credits}
\label{sec:image_credits}
We include the links to the images we used in Figure~\ref{fig:samples} in Table~\ref{tab:image_credits_fig_2}.

\end{document}